\documentclass[twoside,11pt]{article}

\usepackage{dndmodified}
\usepackage{times}
\usepackage{multirow}
\usepackage{stmaryrd}
\usepackage{hyperref}

\usepackage{color}

\newcommand{\dstcslot}[1] {\textrm{#1}}

\usepackage{bbold}

\newcommand{\FunName}[1]{\mathrm{#1}}
\newcommand{\Other}[1]{#1'}
\newcommand{\OOther}[1]{#1''}

\newcommand{\StringSet}[0]{{\mbox{\tt string}}}
\newcommand{\String}[1]{\mbox{\tt "#1"}}

\newcommand{\Ith}[2]{#1_{#2}}
\newcommand{\Set}[1]{\left\{ {#1} \right\}}

\newcommand{\Sets}[2]{{\left\{ {#1}\;\middle\vert\; {#2}\right\}}}

\newcommand{\SpanSet}[4]{{\Set{\Ith {#1} {#2}, \cdots, \Ith {#1} {#3} , \cdots, \Ith {#1} {#4}}}}

\newcommand{\PowerSet}[1]{{{#1}_\subset}}
\newcommand{\SingleSet}[1]{{{#1}_{\Set 1}}}

\newcommand{\Card}[1]{\left| {#1} \right|}
\newcommand{\Cart}[0]{\times}
\newcommand{\Union}[0]{\cup}

\newcommand{\UnionIter}[01]{\displaystyle\bigcup_{#1}}

\newcommand{\And}[0]{\mbox{ and }}
\newcommand{\Or}[0]{\mbox{ or }}
\newcommand{\Indic}[1]{{{\mathbb 1}_{#1}}}

\newcommand{\FunSet}[2]{{{#2}^{#1}}}
\newcommand{\Fx}  [2]{{#1} \left( #2         \right)}
\newcommand{\Fxx} [3]{{#1} \left( #2, #3     \right)}
\newcommand{\Fxxx}[4]{{#1} \left( #2, #3, #4 \right)}

\newcommand{\Pair}[2]{{\left({#1},{#2}\right)}}
\newcommand{\Group}[1]{{\left({#1}\right)}}

\newcommand{\SlotSet}[0]{{\cal S}}

\newcommand{\Slot}[0]{s}

\newcommand{\Value}[0]{v}
\newcommand{\VValue}[0]{{\Other \Value}}
\newcommand{\Valuei}[1]{{\Value}_{#1}}

\newcommand{\Goal}[0]{g}
\newcommand{\GGoal}[0]{\Other \Goal}
\newcommand{\GoalSet}[0]{{\cal G}}

\newcommand{\Hyp}[0]{h}

\newcommand{\HypSet}[0]{{\cal H}}

\newcommand{\Info}[0]{i}

\newcommand{\InfoSet}[0]{{\cal I}}

\newcommand{\Belief}[0]{b}

\newcommand{\BeliefSet}[0]{{\cal B}}

\newcommand{\ValuesOf}[0]{{\FunName{val}}}
\newcommand{\Ukn}[0]{\mbox{\tt ?}} % unknown
\newcommand{\Dc}[0]{\mbox{\tt *}} % don't care
\newcommand{\Not}[0]{\neg} % not
\newcommand{\SetDc}[1]{{#1}_{\Dc}}
\newcommand{\SetUkn}[1]{{#1}_{\Ukn}}
\newcommand{\SetGoal}[1]{{#1}_{\Dc\Ukn}}
\newcommand{\SetNot}[1]{\overline{#1}}

\newcommand{\MachineActLabels}[0]{{\mathbb M}}
\newcommand{\UserActLabels}[0]{{\mathbb U}}
\newcommand{\MachineActs}[0]{{\cal M}}
\newcommand{\UserActs}[0]{{\cal U}}
\newcommand{\MachineUtterance}[0]{m}
\newcommand{\UserUtterance}[0]{u}
\newcommand{\UUserUtterance}[0]{\Other \UserUtterance}
\newcommand{\UUUserUtterance}[0]{\OOther \UserUtterance}

\newcommand{\ProbaSet}[1]{{\widetilde{#1}}}

\newcommand{\Turn}[0]{t}

\newcommand{\pTurn}[0]{{\Turn-1}}
\newcommand{\AtTurn}[2]{#1_{#2}}

\newcommand{\This}{{\String{this}}}
\newcommand{\DontCare}{\Dc}%{\String{dontcare}}}

\newcommand{\ActFormat}[1]{\mbox{\sc #1}}
\newcommand{\SlotFormat}[1]{\mathbf{#1}}
\newcommand{\ValueFormat}[1]{#1}

\newcommand{\SVPair}[2]  {{\SlotFormat{#1} \mbox{\tt =} \ValueFormat{#2}}}
\newcommand{\SVDcPair}[1]{{\SlotFormat{#1} \mbox{\tt =} \Dc}}
\newcommand{\Actx}[2]{{\Fx {\ActFormat{#1}}{#2}}}

\newcommand{\Act}[1]{\Actx {#1} {}}
\newcommand{\Actp}[3]{{\Actx {#1} {\SVPair {#2}{#3}}}} 
\newcommand{\Actpdc}[2]{{\Actx {#1} {\SVDcPair {#2}}}}
\newcommand{\Actpp}[5]{{\Actx {#1} {\SVPair {#2}{#3}, \SVPair {#4}{#5}}}}

\newcommand{\ElemUpdate}[0]{\mu}
\newcommand{\Update}[0]{\tau}
\newcommand{\Extract}[0]{\xi}
\newcommand{\Unthis}[0]{\rho}
\newcommand{\Unthisut}[0]{\varphi}
\newcommand{\Unthishyp}[0]{\FunName{deref}}

\newcommand{\Infmus}[0]{{\FunName{inf}_{\MachineUtterance,\UserUtterance}^\Slot}}
\newcommand{\Infmu}[1]{{\FunName{inf}_{\MachineUtterance,\UserUtterance}^{#1}}}

\newcommand{\Infpmus}[0]{{\FunName{pos}_{\MachineUtterance,\UserUtterance}^\Slot}}
\newcommand{\Infpmu}[1]{{\FunName{pos}_{\MachineUtterance,\UserUtterance}^{#1}}}
\newcommand{\Infmmus}[0]{{\FunName{neg}_{\MachineUtterance,\UserUtterance}^\Slot}}
\newcommand{\Infmmu}[1]{{\FunName{neg}_{\MachineUtterance,\UserUtterance}^{#1}}}

\newcommand{\TabIO}[2]{{#1}\shortrightarrow{#2}}
\newcommand{\TabFun}[2]{{\left[#1,\TabIO \bullet {#2}\right]}}
\newcommand{\BeginTabFun}[0]{\left[\begin{array}{rl}}
\newcommand{\tabIO}[2]{{#1}\shortrightarrow & {#2}\\}
\newcommand{\EndTabFun}[1]{\bullet\shortrightarrow & {#1}\end{array}\right]}

\jmlrheading{vv}{2015}{pp}{xx/xx}{yy/yy}{zz/zz}{Jeremy Fix, Herv{\'e} Frezza-Buet}{doi}

\ShortHeadings{Yet Another Rule Based belief Update System}{Fix and Frezza-Buet}
\firstpageno{1}

\begin{document}

\title{YARBUS : Yet Another Rule Based belief Update System}

\author{\name J{\'e}r{\'e}my Fix \email jeremy.fix@centralesupelec.fr \\
       \addr IMS - MaLIS \& UMI 2958 (GeorgiaTech - CNRS)\\
       Centale-Sup{\'e}lec, 2 rue Edouard Belin\\
       57070 Metz, France
       \AND
       \name Herv{\'e} Frezza-Buet \email herve.frezza-buet@centralesupelec.fr \\
       \addr IMS - MaLIS \& UMI 2958 (GeorgiaTech - CNRS)\\
             Centale-Sup{\'e}lec, 2 rue Edouard Belin\\
             57070 Metz, France
}

%\editor{Jason D. Williams, Antoine Raux, Matthew Henderson}
\editor{}

\maketitle

\begin{abstract}%
  We introduce a new rule based system for belief tracking in dialog systems. Despite the simplicity of the rules being considered, the proposed belief tracker ranks favourably compared to the previous submissions on the second and third Dialog State Tracking challenges. The results of this simple tracker allows to reconsider the performances of previous submissions using more elaborate techniques. 
\end{abstract}


\section{Introduction}

Spoken dialog systems are concerned with producing algorithms allowing a user to interact in natural language with a machine. Particularly important tasks of spoken dialog systems are form-filling tasks~\citep{Goddeau1996}. Booking a flight ticket, querying a restaurant proposing a specific kind of food, looking for a bus from one destination to another can all be framed as a form-filling task. In these tasks, a predetermined and fixed set of \dstcslot{slots} is filled by the machine as the user interacts with it. The form is actually invisible to the user and allows the machine to bias the direction of the dialog in order to uncover the intents of the user. For example, to correctly indicate a bus from one destination to another, the machine has to obtain all the relevant informations such as the source and the destination as well as the time schedule. A typical dialog system is a loop connecting the speech of the user to the spoken synthesis of the machine reply~\citep{Rieser2011}, with a pipeline of processes. Several modules come into play in between these two spoken utterances. The utterance of the user is first processed by an Automatic Speech Recognizer (ASR), feeding a Spoken Language Understanding (SLU) module which outputs a probability distribution over so-called dialog acts. Sequences of dialog acts are handful representations of the intention of a user. These are then integrated by the dialog manager which is in charge of producing the dialog act of the machine which is converted to text and synthesized. There are various sources of noise that can impair the course of the dialog. One of interest for the following of the paper is the noise produced when recognizing what the user said and transcribing it into sequences of dialog acts. Typical semantic parsers (SLU) therefore produce a probability distribution over sequences of dialog acts that reflect what the user might have said. In this paper we focus on integrating these hypotheses in order to infer what the user goal is. This task is the belief tracking, a subpart of the dialog manager.

As any machine learning implementation, evaluating the performances of an algorithm requires data. The recently proposed dialog state tracking challenges offer the opportunity to test belief tracking algorithms on common data~\citep{Black2010,Williams2013,Henderson2014a}. The first challenge focused on form-filling, the second added the possibility of goal changes (especially when the constraints provided by the user are too restrictive to bring any possibility) and the third challenge added the difficulty that there are just few labeled data. In this paper, we work on the datasets for the second and third challenges that focus on the same domain of finding a restaurant by providing constraints on their location, food type, name and price range.

There has been a variety of methods for inferring and tracking the goal of the user submitted to this challenge. Some these methods directly work with the live SLU provided in the dataset (e.g. the focus baseline or the Heriot-Watt tracker \citep{ZhuoranLemon2013}). It turns out that the live SLU is of a rather poor quality and some authors suggested alternative semantic parsers \citep{Sun2014b,Williams2014} or trackers working directly from the ASR output \citep{Henderson2014b}. As a dialog is performed turn by turn, belief tracking can be formulated as an iterative process in which new evidences provided by the SLU are integrated with the previous belief to produce the new belief. In \citet{Sun2014a}, the authors consider the slots to be independent and learn an update rule of the marginal distributions over the goals. The rule they train, taking as one of the inputs the previous belief, is a polynomial function of the probabilities that the user informed or denied a value for a slot, informed or denied a value different from the one for which the output is computed. As the size of hypothesis space grows exponentially with the degree of the polynomial, constraints are introduced in order to prune the space of explored models and to render tractable the optimization problem.  In \citet{Henderson2014b}, the authors explore the ability of recurrent neural networks to solve the belief tracking problem by taking directly as input the speech recognizer output; it does not require any semantic parser. Their work benefits from the recent developments in deep learning. The number of parameters to learn is so large that if the network is not trained carefully, it would not be able to perform well. As shown by the authors, the recurrent neural network performs well on the dataset and their sensitivity to the history of the inputs certainly contribute to their performance. \citet{Williams2014} brought several contributions. The first is the proposition of building up multiple SLU to feed a belief tracker. The second relies in identifying the problem of belief tracking with the problem of the ranking of relevance of answers (documents) to queries which leads the author to propose an interesting approach to the scoring of joint goals. Similarly to document ranking, features (around 2000 to 3000) are extracted from the SLU hypotheses and machine utterance and a regressor is trained to score the different joint goals accumulated so far in the dialog. The tracker proposed by \citet{Williams2014} ranked first at the time of the challenge evaluation. 

One of the attractiveness of the last two methods is their ability to solve the belief tracking problem without requiring much of expert knowledge in dialog systems. These methods extract a medium to large set of features feeding a regressor trained on the datasets. However, there is one potential caveat in these methods which comes from their black-box approach. Indeed, the results of the authors certainly show that the method consisting in exploding the number of attributes extracted from a turn (or previous turns as well) and then training a regressor on this large set of features performs favourably. Nevertheless, at the same time, it tends to lose the grip on the very nature of the data that are processed. As we shall see in this paper, the very limited set of rules employed in YARBUS is extremely simple, yet effective. The paper is organized as follows. Section~\ref{sec:methods} presents all the steps as well as the rules employed in the YARBUS belief tracker. The results of the YARBUS tracker on the DSTC2 and DSTC3 challenges are given in section~\ref{sec:results} and a discussion concludes the paper. The source code used to produce the results of the paper are shared online\citep{yarbussrc}.

\section{Methods}
\label{sec:methods}

The belief tracker we propose computes a probability distribution over the joint goals iteratively. At each turn of a dialog, the utterance of the user is processed by the Automatic Speech Recognized (ASR) module which inputs the semantic parser (SLU) which outputs a probability distribution over dialog acts, called the SLU hypotheses in the rest of the paper. In this paper, three SLU are considered: the live SLU provided originally in the dataset as well as the SLU proposed in \citep{Sun2014b} for the DSTC2 challenge and in \citep{Zhu2014} for the DSTC3 challenge. YARBUS proceeds as following: some pre-processing are operated on the machine acts (getting rid of $\Act{repeat}$ acts) and on the SLU hypotheses (solving the reference of $\String{this}$), then informations are extracted from these reprocessed hypotheses and the belief is updated. Before explaining in details all these steps in the next sections, we introduce some handful notations. 

\subsection{Dialog State Tracking Challenge datasets}

In the following of the paper, we focus on the datasets provided by the Dialog State Tracking Challenges 2 and 3. These contain labeled dialogs for the form-filling task of finding a restaurant. In this section, we provide the keypoints about these datasets and a full detailed description and the data can be found in \citet{dstchandbook}. There are four slots to fill in the DSTC-2 challenge\footnote{the count of values for each slot takes into account the special value $\String{dontcare}$}~: $\SlotFormat {area}$ ($6$ values), $\SlotFormat{{name}}$ ($114$ values), $\SlotFormat{food}$ ($92$ values) and $\SlotFormat{pricerange}$ ($4$ values). This leads to a joint goal space of $374325$ elements including the fact one slot might be unknown. The joint goal space is significantly larger in the third challenge. Indeed, in the third challenge, there are $9$ slots~:$\SlotFormat{area}$ ($16$ values),$\SlotFormat{{chilren allowed}}$ ($3$ values), $\SlotFormat{food}$ ($29$ values), $\SlotFormat{{has internet}}$($3$ values), $\SlotFormat{{has tv}}$ ($3$ values), $\SlotFormat{name}$ ($164$ values), $\SlotFormat{near}$ ($53$ values), $\SlotFormat{pricerange}$ ($5$ values) and $\SlotFormat{type}$ ($4$ values). This leads to a joint goal space of more than $8. 10^9$ elements.

The DSTC-2 challenge contains three datasets: a training set (dstc2\_train) of 1612 dialogs with a total of 11677 turns, a development set (dstc2\_dev) of 506 dialogs with 3934 turns and a test set (dstc2\_test) of 1117 dialogs with 9890 turns. At the time of the challenge, the labeled of only the two first sets were released but we now have the labels for the third subset. The DSTC-3 challenge, which addressed the question of belief tracking when just few labeled data were available and also used a larger set of informable slots, contains two subsets : a training subset (dstc3\_seed) of 10 labeled dialogs with 109 turns and a test set (dstc3\_test) of 2264 dialogs with 18715 turns. The DSTC-2 challenge data contain dialogs in which the machines was driven by one out of three dialog managers in two acoustic conditions and the data of the third challenge were collected with one out of four dialog managers, all in the same acoustic conditions.\\

\subsection{Notations}

The elements handeled by the belief tracking process are strings. Let us denote the set of strings by $\StringSet$. For a set $A$, let us also define $\PowerSet A$ as the set of all the {\em finite} subsets of $A$ (this is a way to represent lists\footnote{E.g. $\PowerSet{\Set{a,b,c}} = \Set{\emptyset, \Set a, \Set b, \Set c, \Set{a,b}, \Set{a,c},\cdots,\Set{a,b,c}}$} with distinct elements) and $\FunSet A B$ the set of the functions from $A$ to $B$. As the context is the filling of the values for the slots in a form, let us denote by $\SlotSet = \SpanSet \Slot 1 2 {\Card \SlotSet}\in \PowerSet \StringSet$ the different slots to be filled. The definition of the slots is part of the ontology related to the dialog. The ontology also defines the acceptable values for each slot. Let us model the slot definition domain thanks to a function $\ValuesOf \in \FunSet \SlotSet {\PowerSet \StringSet}$ which defines $\Fx \ValuesOf \Slot = \Set{\Valuei 1, \Valuei 2, \cdots, \Valuei {n_\Slot}}$ as the set of acceptable values for the slot $\Slot$. Let us consider two extra slot values $\Dc$ and $\Ukn$, respectively meaning that the user does not care about the value of the slot and that the system does not know what the user wants for this slot. 

Let us call a {\em goal} the status of a form, where a slot can be informed or not, depending on what has been said during the interaction with the user. For each slot $\Slot$, a goal specifies a value in $\Fx \ValuesOf \Slot \Union \Set \Dc$ if the slot is informed in the goal, or $\Ukn$ if it has not been informed yet. Using the notations $\SetDc A = A \Union \Set{\Dc}$, $\SetUkn A = A \Union \Set{\Ukn}$, $\SetGoal A = A \Union \Set{\Dc,\Ukn}$, a goal can be defined as $\Goal \in \GoalSet$, with
\begin{equation}
\GoalSet = \Sets{ \Goal \in \FunSet \SlotSet {\Group{\SetGoal \StringSet}}}{\forall \Slot \in \SlotSet,\; \Fx \Goal \Slot \in \SetGoal{\Fx \ValuesOf \Slot}}
\end{equation}
Let us now formalize utterances. Utterances are made of dialog acts, that may differ according to the speaker (user or machine). Acts are coded as a label and a set of slot-value pairs. The labels for machine acts are denoted by $\MachineActLabels = \Set{\String{affirm},\String{bye},\String{canthear}, \cdots}$ and the ones for user acts by $\UserActLabels = \Set{\String{ack},\String{affirm},\String{bye}, \cdots}$. 
For example $\Act{ack}$, 
$\Actpdc{inform}{food}$ 
and $\Actpp{canthelp}{food}{british}{area}{south}$ 
are dialog acts\footnote{For the sake of clarity, a slot-value pair $\Pair{\String a}{\String b}$ is denoted by $\SVPair a b$ and a dialog act $\Pair{\String {act}}{\Set{\Pair{\String {slot}}{\String {value}},\Pair{\String{foo}}{\String{bar}}}}$ by $\Actpp {act}{slot}{value}{foo}{bar}$}.
Let the machine acts be the elements in
\begin{equation}
\MachineActs = \Sets{
\Pair {a} {args} \in \MachineActLabels \times \PowerSet{\Group{\SlotSet \times \SetDc{\StringSet}}}
 }{\forall \Pair \Slot \Value \in args,\; \Value \in \SetDc{\Fx \ValuesOf \Slot}}
\end{equation} 
Let us define user acts $\UserActs$ similarly, using $\UserActLabels$ instead of $\MachineActLabels$. 
\begin{equation}
\UserActs = \Sets{
\Pair {a} {args} \in \UserActLabels \times \PowerSet{\Group{\SlotSet \times \SetDc{\StringSet}}}
 }{\forall \Pair \Slot \Value \in args,\; \Value \in \SetDc{\Fx \ValuesOf \Slot}}
\end{equation} 

We can thus denote a machine utterance by $\MachineUtterance \in \PowerSet{\MachineActs}$ and a user utterance by $\UserUtterance \in \PowerSet{\UserActs}$. The SLU hypotheses are a set of user utterances with a probability for each one. Let us use the notation\footnote{The definition stands for finite sets.} $\ProbaSet A = \Sets{f \in \FunSet A {[0,1]}}{\sum_{a \in A} \Fx f a = 1}$ for the definition of the SLU hypotheses space $\HypSet = \ProbaSet{\PowerSet{\UserActs}}$. An SLU hypothesis is thus denoted by $\Hyp \in \HypSet$.

The principle of the rule based belief tracker presented in this paper is to handle a distribution probability $\Belief \in \BeliefSet = \ProbaSet \GoalSet$ that is updated at each dialog turn $\Turn$. The update (or transition) function $\Update \in \FunSet {\BeliefSet  \Cart \PowerSet{\MachineActs} \Cart \HypSet} \BeliefSet$ is used as
\begin{equation}
\AtTurn \Belief \Turn = \Fxxx \Update {\AtTurn \Belief \pTurn} {\AtTurn \MachineUtterance \Turn} {\AtTurn \Hyp \Turn}
\end{equation}

As detailed further, the update $\tau$ is based on a process that extracts informations from the current SLU hypotheses $\AtTurn \Hyp \Turn$ and the current machine utterance $\AtTurn \MachineUtterance \Turn$. This process consists formally in extracting informations from every user utterances $\UserUtterance \in \PowerSet \UserActs$. However, in the implementation, this combinatorial extraction is avoided by taking the probability $\Fx \Hyp \UserUtterance$ into account and therefore ignoring the huge amounts of null ones. Let us define here what an information about a dialog turn is. Let us use $\SetNot A = \Sets{\Not a}{a \in A}$ the set of the negated values of $A$. An information $\Info \in \InfoSet$ is a function that associates for each slot a value that can be a string, $\Dc$ if the user does not care about the slot value or $\Ukn$ if nothing is known about what the user wants for that slot. An information can also be a set of negated values ($\Dc$ can be negated as well), telling that it is known that the user does not want any of the values negated in that set. This leads to the following definition for $\InfoSet$. 
\begin{eqnarray}
V &=& \SingleSet{\Group{\SetGoal \StringSet}} \Union \PowerSet{\Group{\SetNot{\SetDc \StringSet}}}\nonumber \\
\InfoSet &=& \Sets {\Info \in \FunSet \SlotSet V}{\forall \Slot \in \SlotSet,\; \Fx \Info \Slot \in \SingleSet{\Group{\SetGoal{\Fx \ValuesOf \Slot}}} \Union
\PowerSet{\Group{\SetNot{\SetDc {\Fx \ValuesOf \Slot}}}}}
\end{eqnarray}
where $\SingleSet A$ denotes $\Sets{\Set a}{a \in A}$. For example, $\Fx \Info {\SlotFormat{area}} = \Set{\ValueFormat {east}}$, $\Fx \Info {\SlotFormat{area}} = \Set\Dc$, $\Fx \Info {\SlotFormat{area}} = \Set\Ukn$, $\Fx \Info {\SlotFormat{area}} = \Set{\Not {\ValueFormat {east}}, \Not \Dc}$ are possible information values for the slot $\SlotFormat{area}$. As previously introduced, our rule-based belief tracking process is based on an information extraction from a machine utterance $\MachineUtterance$ and some consecutive user utterance $\UserUtterance$, the probabilities of the SLU hypotheses being handled afterwards. Let us denote this process by a function $\Extract \in \FunSet{{\PowerSet \MachineActs}\Cart{\PowerSet \UserActs}} {\PowerSet\InfoSet}$ such that $\Info \in \Fxx \Extract \MachineUtterance \UserUtterance$ is one instance of all the information extracted by $\Extract$ from $\MachineUtterance$ and $\UserUtterance$.

\subsection{Preprocessing the machine acts and SLU hypotheses}

In order to process the utterances of the user in the correct context, any occurrence of the $\ActFormat{repeat}$ machine act is replaced by the machine act of the previous turn. In the formalization of the user acts, there is one ambiguity that must be solved: some acts contain a $\SlotFormat{this}$ in their slots. In the DSTC challenge, it can occur only for an $\ActFormat{inform}$ act as a $\Actpdc{inform}{this}$. In Yarbus, the attempt to solve the reference of the slot $\SlotFormat{this}$ is based on the occurence of machine acts that explicitely require the user to mention a slot, namely the $\ActFormat{request}$, $\ActFormat{expl-conf}$ and $\ActFormat{select}$ acts. Therefore, the first step is to build up the set $S_\MachineUtterance$ of the slots associated with such acts in the machine utterance~:
\begin{eqnarray*}
A &=& \UnionIter{\Sets{\Pair {a} {args} \in \MachineUtterance}{a \in \Set{\String{expl-conf},\String{select}}}} \UnionIter{\Pair \Slot \Value \in args} \Set\Slot \\
B &=& \UnionIter{\Sets{\Pair {a} {args} \in \MachineUtterance}{a = \String{request}}} \UnionIter{\Pair \Slot \Value \in args} \Set\Value \\
S_\MachineUtterance &=& A \Union B
\end{eqnarray*}
The set $S_\MachineUtterance$ can then be used to rewrite a single user act which can be formally defined by equation~(\ref{eq:this_user_act}).
\begin{equation}
\Unthis \in \FunSet{\UserActs\Cart\PowerSet\SlotSet}{\PowerSet\UserActs}, \;  \Fxx \Unthis {\Pair {a} {args}} {S_\MachineUtterance} = 
\left\{\begin{array}{ll}
\emptyset & \mbox{if } \Pair \This \DontCare \notin args \Or \Card{S_\MachineUtterance}\neq 1\\
\Set{\Pair {a} {\Pair \Slot \Dc}} & \mbox{otherwise, with } S_\MachineUtterance = \Set\Slot
\end{array}\right.
\label{eq:this_user_act}
\end{equation}
As can be seen by the above definition, the result of rewriting a user act is a set with one or no element. The formal definitions of rewriting the SLU is actually easier is rewriting a single dialog act results as a set. The result is an empty set if there is no such $\SVDcPair{this}$ in the slot value pairs of the user act or if there is more than one candidate for the reference. Processing a user utterance (a collection of acts), defined by equation~(\ref{eq:this_user_utterance}), consists in building up the set of all acts that do not contain a $\SVDcPair{this}$ slot-value pair and then complement it with the rewritten acts.

\begin{equation}
\begin{array}{lll}
\Unthisut \in \FunSet{\PowerSet\UserActs\Cart\PowerSet\MachineActs}{\PowerSet\UserActs}, \; \Fxx \Unthisut \UserUtterance \MachineUtterance 
= & & \UserUtterance \setminus \Sets{\Pair {a} {args} \in \UserUtterance}{\Pair \This \DontCare \in args} \\
 & \Union & \UnionIter{\UUserUtterance \in \UserUtterance} \Fxx \Unthis \UUserUtterance {S_\MachineUtterance}
\end{array}
\label{eq:this_user_utterance}
\end{equation}
As dropping acts for the user utterance can result in creating duplicate hypotheses in the SLU, the final step consists in merging these duplicates and summing their probabilities~:
\begin{equation}
\Unthishyp \in \FunSet{\HypSet\Cart\PowerSet\MachineActs}{\HypSet},\;
\Fx{\Fxx \Unthishyp \Hyp \MachineUtterance}{\UserUtterance} = \sum_{\UUserUtterance \in \Sets{\UUUserUtterance \in \PowerSet\UserActs}{\Fxx \Unthisut \UUUserUtterance \MachineUtterance = \UserUtterance}} \Fx \Hyp \UUserUtterance
\end{equation}

The turns in the following table illustrate the SLU hypotheses when trying to solve the reference of $\This$in two situations\footnote{The sentences come from the session-id \textit{voip-db80a9e6df-20130328\_234234} of the DSTC2 test set}. In the turn of the first dialog, the reference is solved with the slot $\SlotFormat{food}$ while it cannot be solved in the turn of the second dialog.

{\footnotesize
\begin{center}
\begin{tabular}{l|lr|lr}
\textbf{System act} & \multicolumn{2}{c|}{\textbf{Original SLU}} & \multicolumn{2}{c}{\textbf{Rewritten SLU}} \\
& Hypothesis & Score & Hypothesis & Score\\
\hline\hline
\multirow{3}{*}{$\Actp{request}
{slot}{food}$}                        & $\Actpdc{inform}{this}$      & 0.99                  & $\Actpdc{inform}{food}$                        & 0.99                  \\
\cline{2-5}
                                      & $\Act{affirm}$               & \multirow{2}{*}{0.01} & $\Act{affirm}$                                 & \multirow{2}{*}{0.01} \\
                                      & $\Actpdc{inform}{this}$      &                       & $\Actpdc{inform}{food}$                        &                       \\
\hline\hline
                                      & $\Actpdc{inform}{this}$      & 0.40                  & \multirow{2}{*}{$\emptyset$}                   & \multirow{2}{*}{0.53} \\
\cline{2-3}
                                      & $\emptyset$                  & 0.13                  &                                                &                       \\
\cline{2-5}
                                      & $\Act{reqalts}$              & 0.14                  & $\Act{reqalts}$ & 0.14  \\
\cline{2-5}
                                      & $\Act{affirm}$               & \multirow{2}{*}{0.13} & \multirow{3}{*}{$\Act{affirm}$}                & \multirow{3}{*}{0.20} \\
                                      & $\Actpdc{inform}{this}$      &                       &                                                &                       \\
 \cline{2-3}
$\Actp{offer}{name}{golden wok}$      & $\Act{affirm}$               & 0.07                  &                                                &                       \\
  \cline{2-5}
$\Actp{inform}{price}{moderate}$      & $\Act{ack}$                  & 0.06                  & $\Act{ack}$  & 0.06 \\
  \cline{2-5}
$\Actp{inform}{area}{north}$          & $\Act{negate}$               & \multirow{2}{*}{0.03} & \multirow{3}{*}{$\Act{negate}$}                & \multirow{3}{*}{0.05} \\
                                      & $\Actpdc{inform}{this}$      &                       &                                                &                       \\
  \cline{2-3}
                                      & $\Act{negate}$               & 0.02                  &                                                &                       \\
  \cline{2-5}
                                      & $\Actpdc{inform}{this}$      & \multirow{2}{*}{0.02} & \multirow{2}{*}{$\Actp{inform}{area}{north}$ } & \multirow{2}{*}{0.02} \\
                                      & $\Actp{inform}{area}{north}$ &                       &                                                &                       \\
  \cline{2-5}
                                      & $\Act{thankyou}$             & 0.01                  & $\Act{thankyou}$ & 0.01  
\end{tabular}
\end{center}}

\subsection{Extracting informations from the SLU hypotheses}
\label{sec:extracting_information}

The rewritten hypotheses resulting from solving the $\This$ reference can now be processed to extract information. Every hypothesis is considered one after the other and a set of basic rules is applied on each. The information extracted from each hypothesis is represented as a tuple with a set of values for each slot. Informally, these rules build up a set for each slot $\Slot$ as~:
\begin{enumerate}
\item If the hypothesis contains a $\Actp{inform}{\Slot}{\Value}$, the information $\Value$ is added to the set,
\item If the hypothesis contains a $\Act{affirm}$, for every $\Actp{expl-conf}{\Slot}{\Value}$ in the machine utterance, the information $\Value$ is added to the set,
\item If the hypothesis contains a $\Actp{deny}{\Slot}{\Value}$, the information $\Not \Value$ is added to the set,
\item If the hypothesis contains a $\Act{negate}$, for every $\Actp{expl-conf}{\Slot}{\Value}$ in the machine utterance, the information $\Not \Value$ is added to the set,
\item If the hypothesis contains no $\Act{negate}$, for every $\Actp{impl-conf}{\Slot}{\Value}$ in the machine utterance, the information $\Value$ is added to the set.
\end{enumerate}
Altogether, these rules capture three ideas. The first is that informed slot-value pairs must be captured wheter positively or negatively depending on the act that inform the slot/value pair. The second is that slot-value pairs that are asked by the machine to be explicitly confirmed must be considered only if the user is explicitly accepting or denying them (in which case the values are integrated positively or negatively) and the last rule integrates information implicitly confirmed by the machine only when the user does not negate them. As we shall see in the result section, this set of simple rules is sufficient to get reasonably good results on the challenges.

The five rules for extracting information from the SLU hypothesis can be formally defined by introducing the sets $\Infpmus$ and $\Infmmus$ in $\PowerSet{\Group{\SetDc\StringSet \Union \SetNot{\SetDc\StringSet}}}$, $\MachineUtterance \in \MachineActs$, $\UserUtterance \in \UserActs$ and $\Slot \in \SlotSet$ as~:
\begin{equation}
\begin{array}{lll}
\Infpmus = & &  \Sets{\Value \in \SetDc{\Fx \ValuesOf \Slot}}{\exists \Pair a {args} \in \UserUtterance,\; a=\String{inform} \And \Pair \Slot \Value \in args}\\

& \Union & \Sets{\Value \in \SetDc{\Fx \ValuesOf \Slot}}{
\begin{array}{ll}
\Pair{\String{affirm}}\emptyset \in \UserUtterance & \\
\mbox{and } \exists \Pair a {args} \in \MachineUtterance, & a=\String{expl-conf} \\ 
& \mbox{and } \Pair \Slot \Value \in args 
\end{array}}\\

& \Union & \Sets{\Value \in \SetDc{\Fx \ValuesOf \Slot}}{
\begin{array}{ll}
\Pair{\String{negate}}\emptyset \notin \UserUtterance & \\
\mbox{and } \exists \Pair a {args} \in \MachineUtterance, & a=\String{impl-conf} \\ 
& \mbox{and } \Pair \Slot \Value \in args 
\end{array}}
\end{array}
\label{eq:pos_mus}
\end{equation}

\begin{equation}
\begin{array}{lll}
\Infmmus = & & \Sets{\Not \Value \in \SetNot{\SetDc{\Fx \ValuesOf \Slot}}}{\exists \Pair a {args} \in \UserUtterance,\; a=\String{deny} \And \Pair \Slot \Value \in args}\\
& \Union & \Sets{\Not \Value \in \SetNot{\SetDc{\Fx \ValuesOf \Slot}}}{
\begin{array}{ll}
\Pair{\String{negate}}\emptyset \in \UserUtterance & \\
\mbox{and } \exists \Pair a {args} \in \MachineUtterance, & a=\String{expl-conf} \\ 
& \mbox{and } \Pair \Slot \Value \in args 
\end{array}}
\end{array}
\label{eq:neg_mus}
\end{equation}

The set $\Infpmus$ (resp. $\Infmmus$) contains the positive (resp. negative) information that can be extracted from the machine utterance and a single user utterance. These two sets are then merged and cleaned. Let us take an example to motivate the cleaning process. Suppose that the user has negated a slot-value pair that the machine requests to explicitly confirm ($\Infmmu{\SlotFormat{food}}=\Set{\Not \ValueFormat{french}}$) and in the same utterance informs s/he has informed that s/he wants a british restaurant ($\Infpmu{\SlotFormat{food}} = \Set{\ValueFormat{british}}$), then the information about the british food is more informative than the $\Not \ValueFormat{french}$ information for uncovering the user's goal. The second motivation comes from possible conflicts. Suppose the machine utterance is $\Actp{expl-conf}{food}{british}$ and that the SLU recognized the utterance $\Act{affirm}\Actp{inform}{food}{french}$. In that case, there is clearly a conflict and there is no \emph{a priori} reason to favor $\SVPair{food}{british}$ over $\SVPair{food}{french}$. In Yarbus, the two extracted positives therefore receive a uniform split of the mass given of the SLU hypothesis from which they are extracted. The step of splitting the mass of an hypothesis over the information extracted from it is made explicit in the next section on the update function. Formally, merging the sets of positives and negatives can be defined as building the set of sets $\Infmus$ as~:
\begin{equation}
\Infmus = \left\{\begin{array}{ll}
\Set\Ukn & \mbox{if } \Pair \Infpmus \Infmmus = \Pair \emptyset \emptyset \Or\UserUtterance = \emptyset \\
\Set\Infmmus & \mbox{if } \Infpmus = \emptyset \\
\begin{array}{l}
\Sets{\Set\Value}{\Value \in \Infpmus} \\
\; \Union \Set{\Sets{\Not \Value \in \Infmmus}{\Value \in \Infpmus}} 
\end{array}& \mbox{otherwise}
\end{array}\right.
\end{equation}
If there is no positive nor negative or if the utterance of the user is empty\footnote{this rule prevents the inclusion of information extracted based on the absence of acts in the user utterance such as the third rule of equation~(\ref{eq:pos_mus}). Indeed, it is more conservative to suppose that an empty utterance might contain the act we are supposing is missing.}, the value for the slot is simply unknown. In case there are no positives, all the negatives are kept. In case there are both positives and negatives, all the positives are kept as singletons as well the negatives that conflict with the positives. An example\footnote{the example is the second turn of \textit{voip-5cf59cc660-20130327\_143457} in dstc2\_test} of information extraction and fusion is given in the following:
\begin{eqnarray}
\MachineUtterance &=& \Actp{expl-conf}{food}{vietnamese} \nonumber \\
\UserUtterance &=& \Set{\Act{negate},\Actpdc{inform}{this},\Actp{inform}{food}{romanian}} \nonumber \\
\Infpmu{\SlotFormat{food}} &=& \Set{\Dc,\ValueFormat{romanian}} \nonumber \\
\Infmmu{\SlotFormat{food}} &=& \Set{\Not \ValueFormat{vietnamese}} \nonumber \\
\Infmu{\SlotFormat{food}} &=& \Set{\Set{\Dc}, \Set{\ValueFormat{romanian}}}
\end{eqnarray}
where both positives and negatives are involved. The positives are retained and the negative is therefore discarded in the fusion.

As Yarbus focuses on joint goals, the cartesian product of the information extracted for each slot is computed and leads to the set of information for all the slots extracted from a single machine utterance and user utterance $\Fxx \Extract \MachineUtterance \UserUtterance$. This can be formally defined as~:
 \begin{equation}
\Fxx \Extract \MachineUtterance \UserUtterance = \Sets{\Info \in \InfoSet}{\forall \Slot, \Fx \Info \Slot \in \Infmus}
\end{equation}

Let us consider for example the machine utterance $\Actp{expl-conf}{pricerange}{cheap}$. The SLU hypothesis $\Hyp \in \HypSet$ as well as their associated information set $\Fxx \Extract \MachineUtterance \UserUtterance, \UserUtterance \in \Hyp$ are shown in the following, where the tabular definition\footnote{this definition is introduced for the sake of clarity of the examples.} of function $\TabFun{\TabIO{x_1}{y_1},\TabIO{x_2}{y_2}}{y_3}$ stands for the function returning $y_1$ for $x_1$,  $y_2$ for $x_2$ and $y_3$ otherwise. Let us consider
\begin{eqnarray}
\MachineUtterance &=& \Set{\Actp{expl-conf}{pricerange}{cheap}} \nonumber\\
\Hyp &=& \BeginTabFun
\tabIO {\Set{\Actpdc{inform}{pricerange}}} {0.87}
\tabIO {\Set{\Act{affirm},\Actpdc{inform}{pricerange}}} {0.10}
\tabIO {\Set{\Act{negate},\Actpdc{inform}{pricerange}}} {0.03}
\EndTabFun {0}\nonumber
\end{eqnarray}
For each hypothesis in $\Hyp$ with a non null probability, the extracted informations are
\begin{eqnarray}
\Fxx \Extract \MachineUtterance {\Set{\Actpdc{inform}{pricerange}}} &=& \Set{\TabFun{\TabIO{\SlotFormat{pricerange}}{\Set\Dc}}{\Set\Ukn}} \nonumber\\
\Fxx \Extract \MachineUtterance {\Set{\Act{affirm},\Actpdc{inform}{pricerange}}} &=& \Set{
\begin{array}{l}
\TabFun{\TabIO{\SlotFormat{pricerange}}{\Set\Dc}}{\Set\Ukn},\\
\TabFun{\TabIO{\SlotFormat{pricerange}}{\Set{\ValueFormat{cheap}}}}{\Set\Ukn}
\end{array}}\nonumber\\
\Fxx \Extract \MachineUtterance {\Set{\Act{negate},\Actpdc{inform}{pricerange}}} &=& \Set{\TabFun{\TabIO{\SlotFormat{pricerange}}{\Set\Dc}}{\Set\Ukn}} \nonumber
\end{eqnarray}

\subsection{Updating the belief from the extracted informations}

The goal of a belief tracker is to update a probability distribution over $\GoalSet$, i.e to update $\AtTurn \Belief \Turn \in \BeliefSet$ at each successive turn $\Turn$. Before the first turn, we assume no \emph{a priori} on the goal of the user and the belief $\AtTurn \Belief 0$ is therefore initialized as~:
\begin{equation}
\Fx{\AtTurn \Belief 0} \Goal = \left\{
\begin{array}{ll}
1 & \mbox{if } \Goal \mbox{ is the constant function } \Fx \Goal \Slot = \Ukn \\
0 & \mbox{otherwise}
\end{array} \right.
\end{equation}

The belief update, denoted by $\Update$, relies on an elementary transition function $\ElemUpdate \in \FunSet{\GoalSet\times\InfoSet}{\GoalSet}$ defined by equation~(\ref{eq:transition_function}).

\begin{equation}
\Fx {\Fxx \ElemUpdate \Goal \Info} \Slot = \left\{\begin{array}{lllll}
\Value & \mbox{if} & \Fx \Goal \Slot =  \Value \in \SetGoal{\Fx \ValuesOf \Slot} & \mbox{and} & \Fx \Info \Slot = \Set \Ukn\\
\Value & \mbox{if} & \Fx \Goal \Slot =  \Ukn & \mbox{and} & \Fx \Info \Slot = \Set \Value,\;\Value \in \SetDc{\Fx \ValuesOf \Slot}\\
\Ukn  & \mbox{if} & \Fx \Goal \Slot =   \Ukn & \mbox{and} & \Fx \Info \Slot \in \PowerSet{\Group{\SetNot{\SetDc{\Fx \ValuesOf \Slot}}}}\\
\VValue & \mbox{if} & \Fx \Goal \Slot =  \Value \in \SetDc{\Fx \ValuesOf \Slot} & \mbox{and} & \Fx \Info \Slot = \Set \VValue,\; \VValue \in \SetDc{\Fx \ValuesOf \Slot}\\
\Ukn & \mbox{if} & \Fx \Goal \Slot =  \Value \in \SetDc{\Fx \ValuesOf \Slot} & \mbox{and} & \Not \Value \in \Fx \Info \Slot \\
\Value & \mbox{if} & \Fx \Goal \Slot =  \Value \in \SetDc{\Fx \ValuesOf \Slot} & \mbox{and} & \Fx \Info \Slot \in \PowerSet{\Group{\SetNot{\SetDc{\Fx \ValuesOf \Slot}} \setminus \Set {\Not \Value}}}\\
\end{array}\right.
\label{eq:transition_function}
\end{equation}
For each slot $\Slot$, the transition function states that the goal $\Value$ remains the same if the information is unknown, $\Value$ or a negation of a value different from $\Value$. The goal changes to unknown in case the information negates it. And finally, if the information is a positive different from the current goal, the goal switches to this positive.

Given the transition function $\ElemUpdate$, the belief can be updated by introducing the belief update function $\Update \in \FunSet {\BeliefSet \times \PowerSet \MachineActs \times \HypSet} {\BeliefSet}$ according to equation~(\ref{eq:belief_update}).
\begin{eqnarray}
P_{\AtTurn \MachineUtterance \Turn,\UserUtterance}^{\GGoal\rightarrow\Goal} &=& \frac 1 {\Card {\Fxx \Extract {\AtTurn \MachineUtterance \Turn} {\UserUtterance}}} \sum_{\Info \in \Fxx \Extract {\AtTurn \MachineUtterance \Turn} {\UserUtterance}} \Fx {\Indic {\Set \Goal}} {\Fxx \ElemUpdate \GGoal \Info}\\
\AtTurn \Belief \Turn = \Fx {\Fxxx \Update {\AtTurn \Belief {\pTurn}} {\AtTurn \MachineUtterance \Turn} {\AtTurn \Hyp \Turn}} \Goal &=&  \sum_{\GGoal \in \GoalSet} \Fx {\AtTurn \Belief {\pTurn}} \GGoal \sum_{\UserUtterance \in \PowerSet \UserActs} \Fx {\AtTurn \Hyp \Turn} \UserUtterance P_{\AtTurn \MachineUtterance \Turn,\UserUtterance}^{\GGoal\rightarrow\Goal} 
\label{eq:belief_update}
\end{eqnarray}
where $P_{\AtTurn \MachineUtterance \Turn,\UserUtterance}^{\GGoal\rightarrow\Goal}$ shares equally the probabilities between all the information extracted from $\AtTurn \MachineUtterance \Turn$ and~$\UserUtterance$, and retains only, once the share is affected, the information generating a transition from $\GGoal$ to $\Goal$.

%% \begin{table}[htbp]
%% \begin{center}
%% \begin{footnotesize}
%% \begin{tabular}{|c|c|c||c|c|c||c|c|c|}
%% \hline
%% \multicolumn{9}{|c|}{DSTC2}\\
%% \multicolumn{3}{|c||}{dstc2\_train} & \multicolumn{3}{c||}{dstc2\_dev} & \multicolumn{3}{c|}{dstc2\_test}\\ 
%% Acc. & L2 & ROC  & Acc. & L2 & ROC  & Acc. & L2 & ROC\\
%% \hline
%% 0.9995616 &0.0008768 &0.0000877 &0.9997393 & 0.0005214& 0.0002608&0.9812158 &0.0394778 &0.0001052\\
%% \hline
%% \end{tabular}
%% \vspace{0.5cm}
%% \begin{tabular}{|c|c|c||c|c|c|}
%% \hline
%% \multicolumn{6}{|c|}{DSTC3}\\
%% \multicolumn{3}{|c||}{dstc3\_seed}&\multicolumn{3}{|c|}{dstc3\_test}\\
%% Acc. & L2 & ROC &Acc. & L2 & ROC\\
%% \hline
%% & & &1.000 & 0.0000849& -\\
%% \hline
%% \end{tabular}
%% \end{footnotesize}
%% \end{center}
%% \caption{\label{table:yarbus_dstc_semantics} \todo{Inutile dans la version finale du papier}Results of Yarbus on the DSTC-2 and DSTC-3 datasets making use of the labeled semantics, i.e. a noise-free condition. For each dataset, the reported scores are the featured metrics of the challenges, namely: ``accuracy'', ``L2 norm'' and ``ROC performance Correct Accept 5\%''.}
%% \end{table}

\section{Results \label{sec:results}}

\subsection{Running the tracker on the noise-free SLU from the labels}

The datasets of the challenges have been labeled using Amazon Mechanical Turk. These labels contain the joint goals that the belief tracker has to identify and also the \emph{semantics}, i.e. what the labelers understood from the audio recordings of the dialogs and written in the dialog acts formalism. Therefore, one can make use of this semantics to test the belief tracker in a noise free SLU condition. In theory, a good belief tracker should have the best scores on the metrics in this ideal condition. It turns out that Yarbus does perform almost perfectly according to the metrics in this ideal condition by performing the following number of mistakes~:
\begin{itemize}
\item 5 mistakes on the joint goals for dstc2\_train
\item 1 mistake on the joint goals for dstc2\_dev
\item 183 mistakes on the joint goals for dstc2\_test
\item 0 mistake on the joint goals for dstc3\_test
\end{itemize}
All the mistakes are actually produced by the $\Act{impl-conf}$ rule and discarding this rule leads to $100\%$ accuracy. Indeed, there are some slot/value pairs that get integrated in the belief by Yarbus because they have been implicitly confirmed by the machine and not denied by the user. This rule is actually not beneficial when the SLU is un-noisy. Indeed, it generates information that is not explicitly given by the user but generated by the machine based on its belief. As we shall see in the next sections, with the outputs from the SLUs, there is a slight improvement in performances if we consider this rule. Actually, we go back on this issue by discussing the performances with respect to the set of rules being considered in section~\ref{subsec:varying_rules}.

\begin{table}[htbp]
\begin{center}
\begin{tabular}{|c|c|c||c|c|c||c|c|c|}
\hline
\multicolumn{9}{|c|}{Live SLU}\\
\multicolumn{3}{|c||}{dstc2\_train} & \multicolumn{3}{c||}{dstc2\_dev} & \multicolumn{3}{c|}{dstc2\_test}\\ 
Accuracy & L2 & ROC  & Accuracy & L2 & ROC  & Accuracy & L2 & ROC  \\
\hline
0.719 & 0.464& 0 &0.630 & 0.602& 0& 0.725& 0.440& 0 \\
\hline\hline
\multicolumn{9}{|c|}{SJTU 1best SLU}\\
\multicolumn{3}{|c||}{dstc2\_train} & \multicolumn{3}{c||}{dstc2\_dev} & \multicolumn{3}{c|}{dstc2\_test}\\ 
Accuracy & L2 & ROC  & Accuracy & L2 & ROC  & Accuracy & L2 & ROC  \\
\hline
0.835 & 0.265& 0.232& 0.801& 0.330& 0.254& 0.752& 0.392&  0.271\\
\hline\hline
\multicolumn{9}{|c|}{SJTU 1best+sys SLU}\\
\multicolumn{3}{|c||}{dstc2\_train} & \multicolumn{3}{c||}{dstc2\_dev} & \multicolumn{3}{c|}{dstc2\_test}\\ 
Accuracy & L2 & ROC  & Accuracy & L2 & ROC  & Accuracy & L2 & ROC  \\
\hline
0.871 & 0.213& 0.281& 0.841& 0.257& 0.208& 0.759& 0.358& 0.329 \\
\hline
\end{tabular}
\end{center}
\caption{\label{table:yarbus_dstc2} Results of Yarbus on the DSTC-2 datasets for the three SLU (live, and the two from SJTU\citep{Sun2014b}). For each dataset, the reported scores are the featured metrics of the challenges, namely: ``accuracy'', ``L2 norm'' and ``ROC performance Correct Accept~5\%''.}
\end{table}

\begin{table}[htbp]
\begin{center}
\begin{tabular}{|c|c|c||c|c|c|}
\hline 
\multicolumn{3}{|c||}{Live SLU} & \multicolumn{3}{c|}{SJTU asr-tied SLU}\\ 
        Acc. & L2 & ROC ca5      & Acc. & L2 & ROC ca5  \\
\hline
 0.582& 0.702& 0&0.597 &0.624 &0.226\\
\hline\hline
\multicolumn{3}{|c||}{SJTU errgen SLU} & \multicolumn{3}{c|}{SJTU errgen+rescore SLU}\\ 
Acc. & L2 & ROC ca5      & Acc. & L2 & ROC ca5  \\
\hline
0.594& 0.624& 0.150& 0.595& 0.607&0.151\\
\hline
\end{tabular}
\end{center}
\caption{\label{table:yarbus_dstc3} Results of Yarbus on the DSTC-3 datasets with four SLU (live, and the three SLU from \citep{Zhu2014}). For each dataset, the reported scores are the featured metrics of the challenges, namely: ``accuracy'', ``L2 norm'' and ``ROC performance Correct Accept~5\%''.}
\end{table}

\subsection{Performances on the challenges and comparison to the previous submissions}

\begin{figure}
\begin{center}
\begin{tabular}{c}
\includegraphics[width=0.7\linewidth]{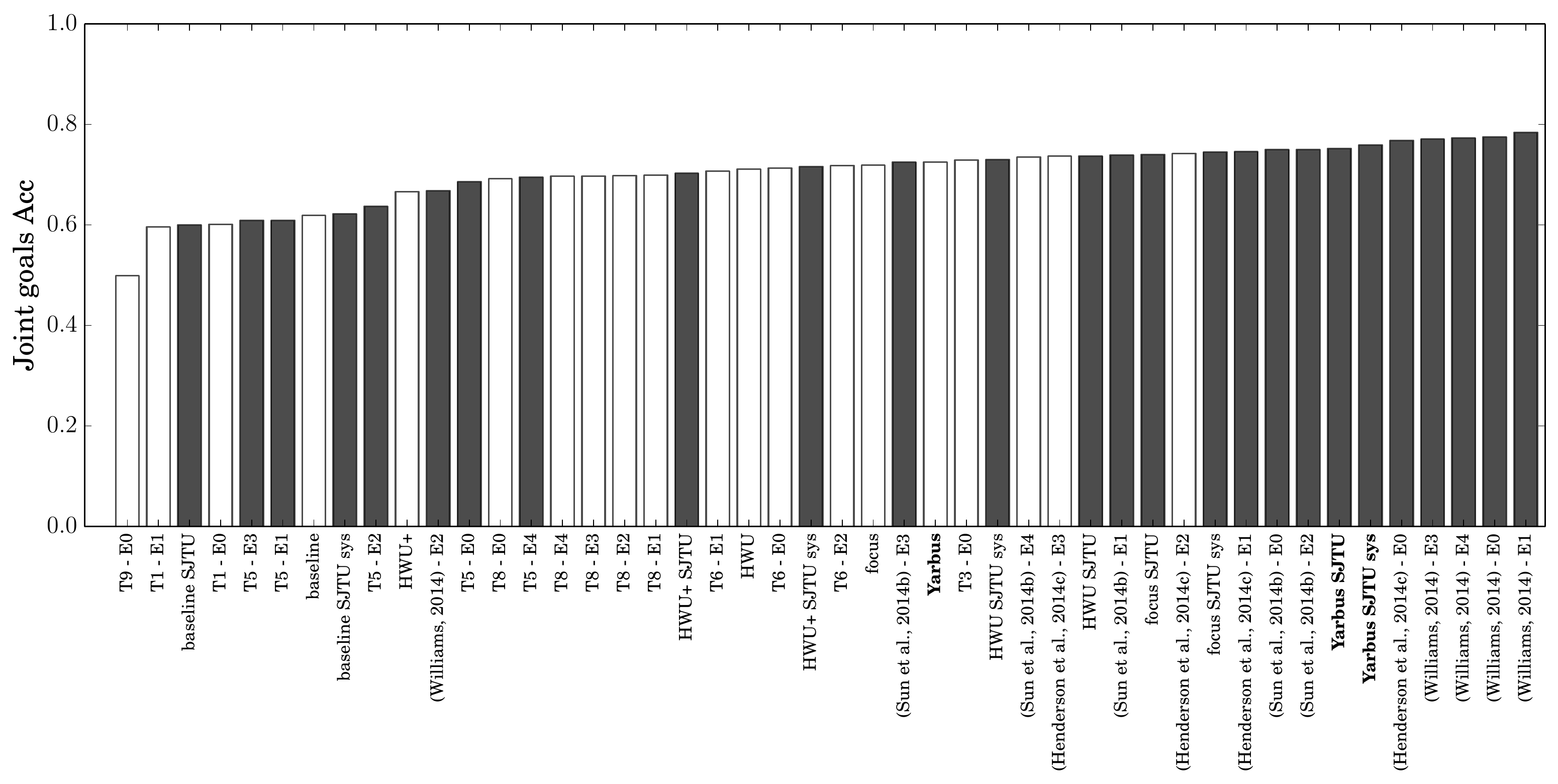}\\
a) \\
\includegraphics[width=0.7\linewidth]{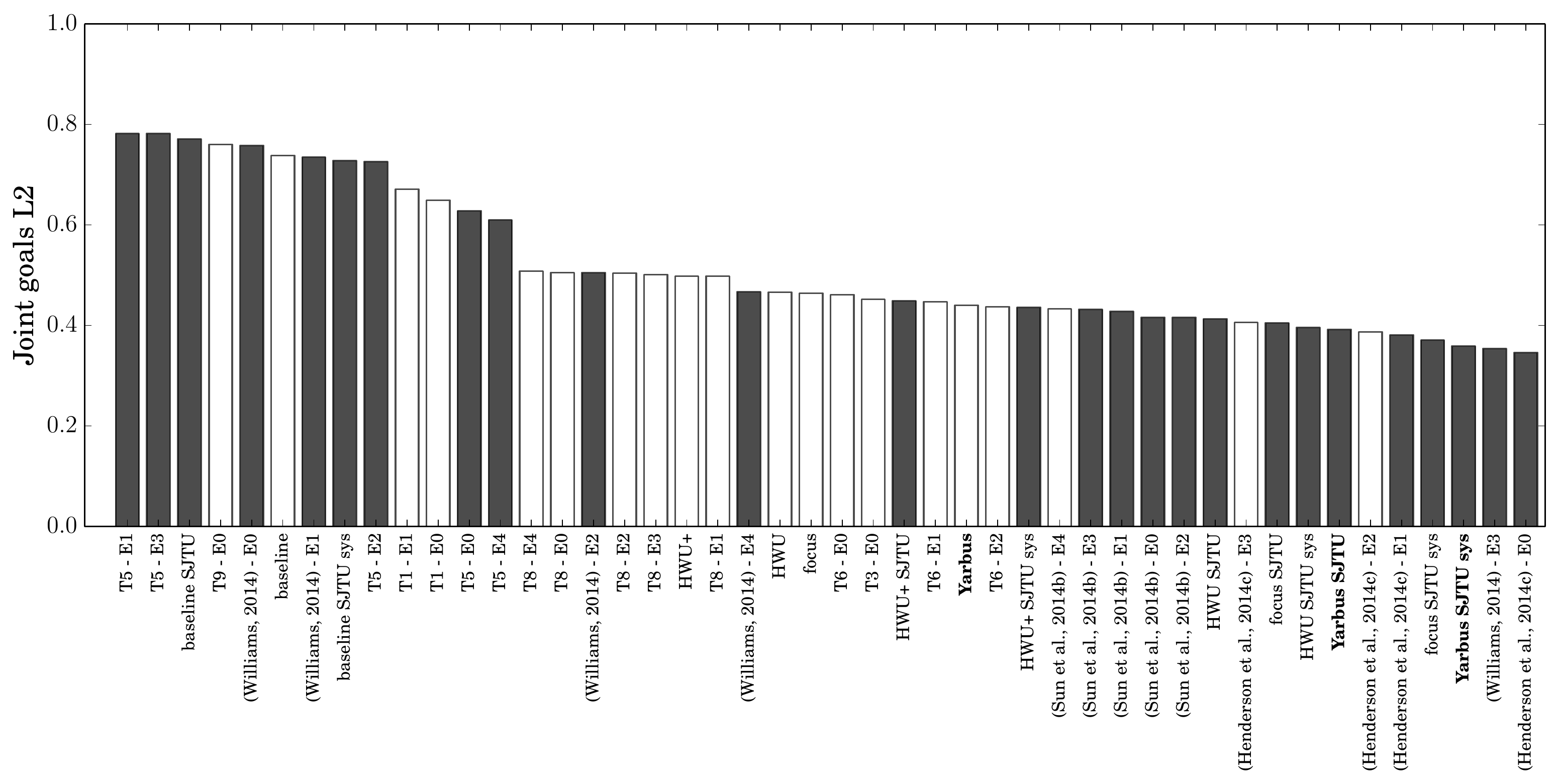}\\
b)  \\
\includegraphics[width=0.7\linewidth]{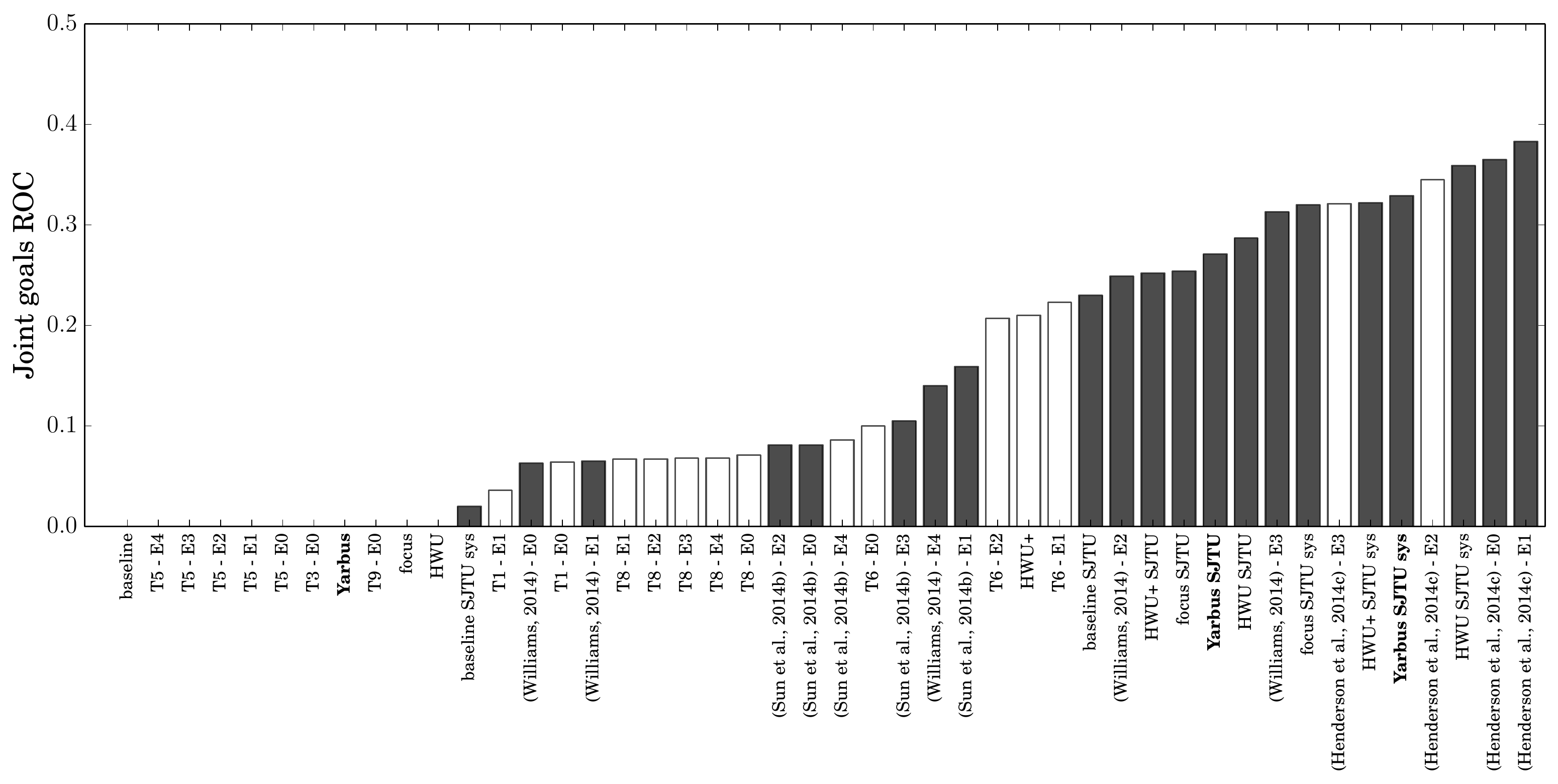}\\
c) 
\end{tabular}
\end{center}
\caption{\label{fig:comparison_dstc2_test}Performances of the trackers on the dstc2\_test dataset. The reported measures are the features metrics of the challenge: a) Accuracy, b) L2 norm, c) ROC CA5\%. The trackers using the live ASR are represented with black bars and the trackers not using the live ASR (i.e. only the live SLU) in white. The y-ranges are adjusted to better appreciate the differences of the scores.}
\end{figure}

\begin{figure}
\begin{center}
\begin{tabular}{c}
\includegraphics[width=0.7\linewidth]{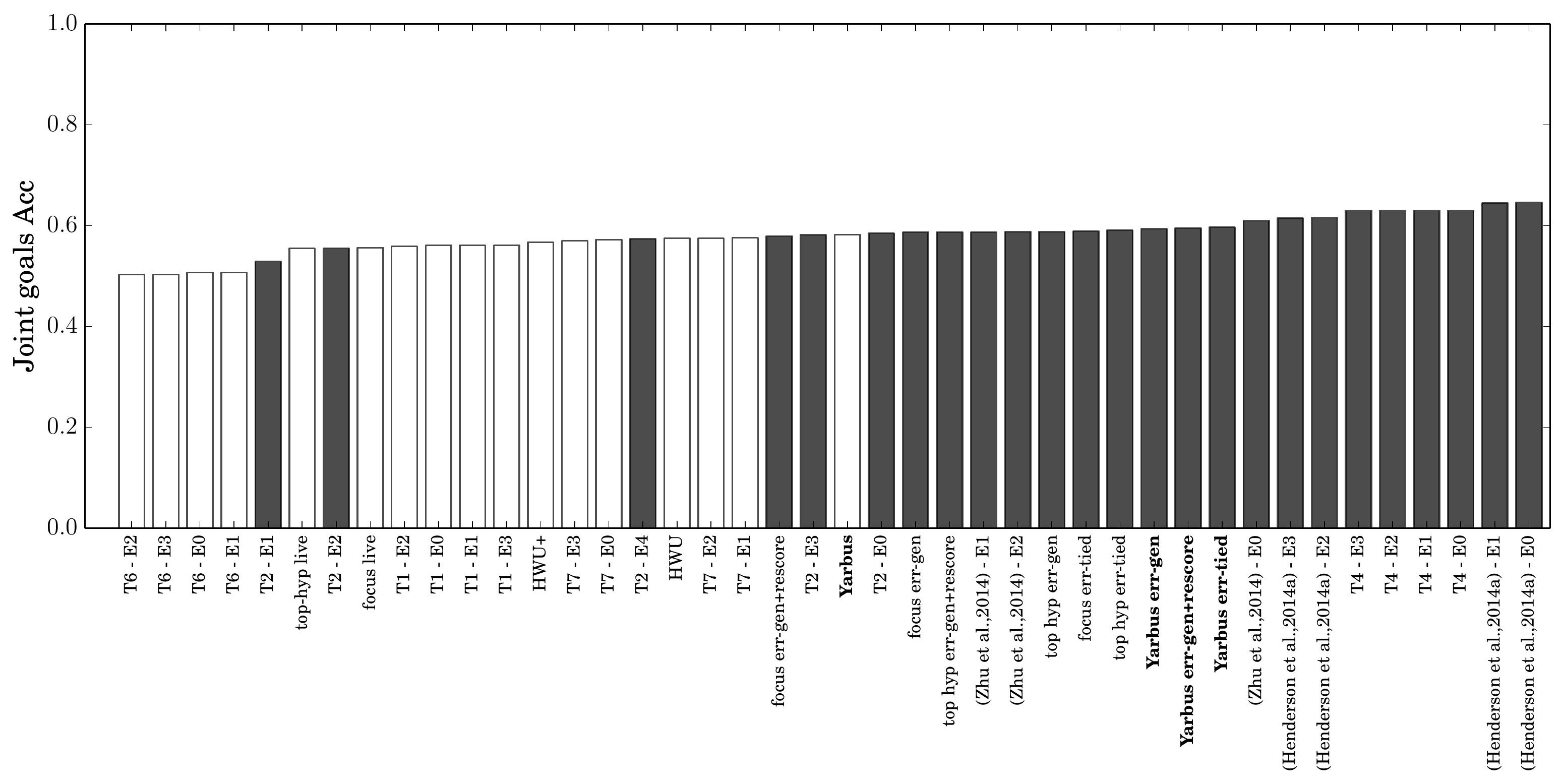}\\
a) \\
\includegraphics[width=0.7\linewidth]{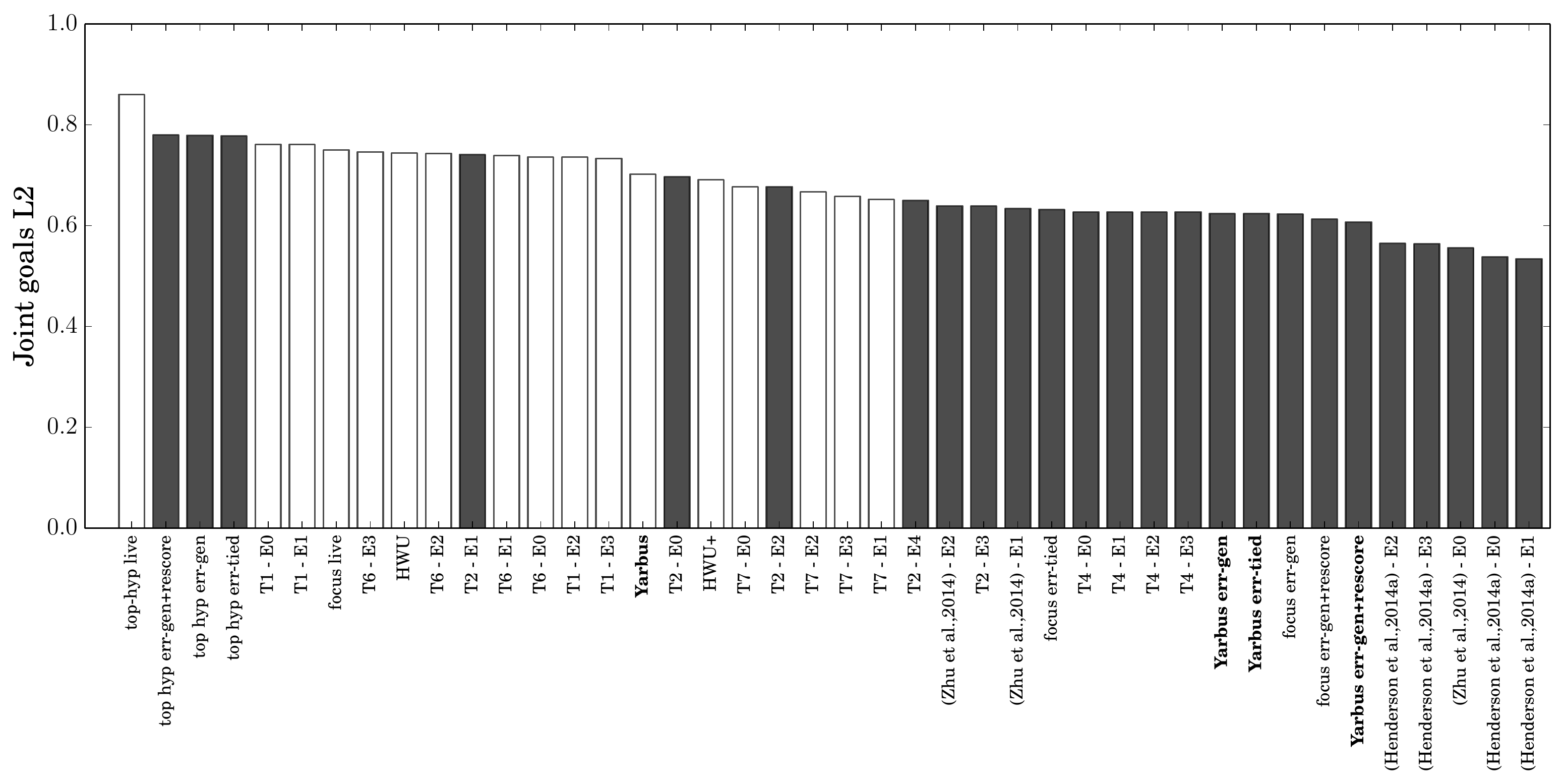}\\
b)  \\
\includegraphics[width=0.7\linewidth]{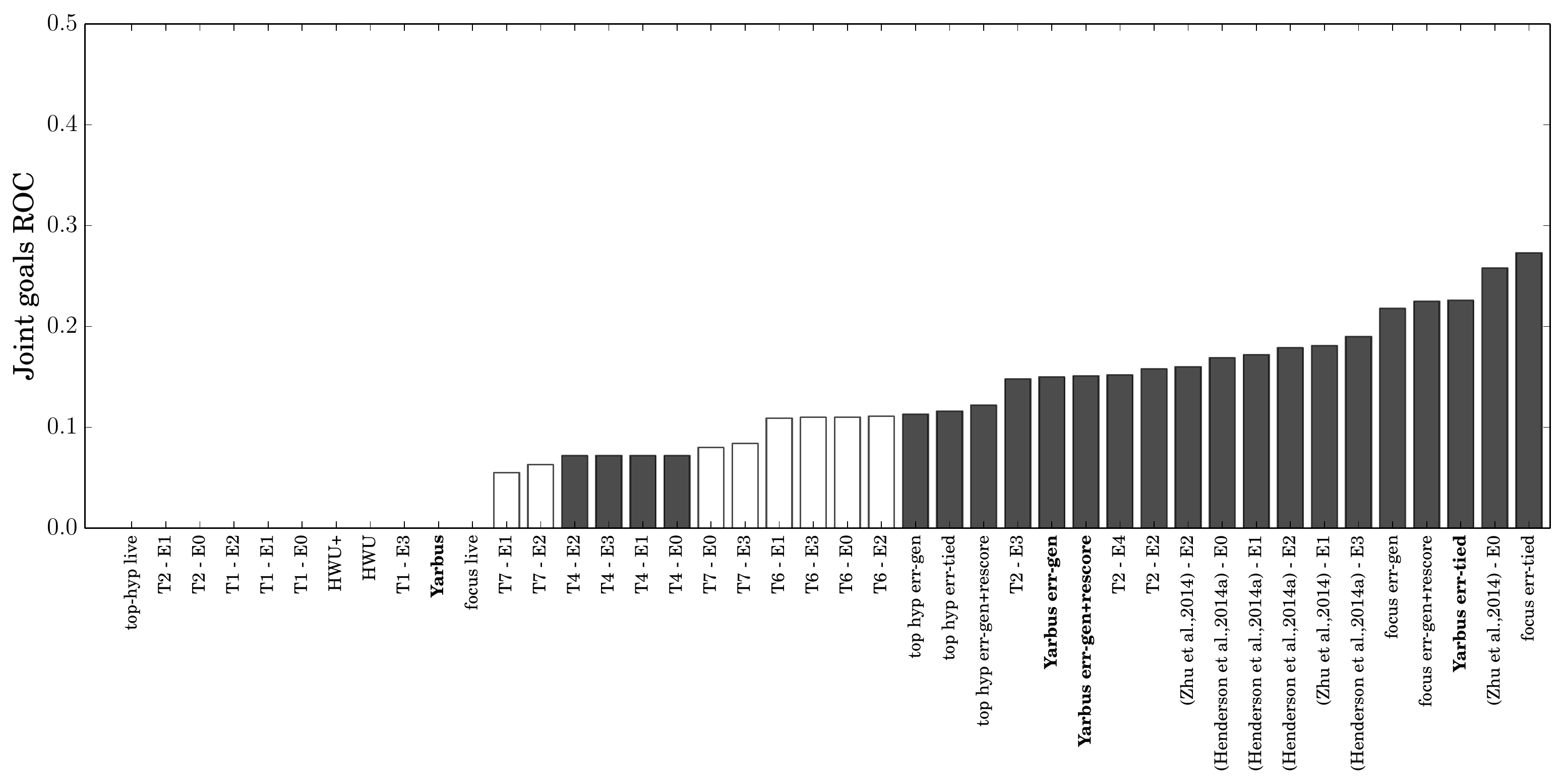}\\
c) 
\end{tabular}
\end{center}
\caption{\label{fig:comparison_dstc3}Performances of the trackers on the dstc3\_test dataset. The reported measures are the features metrics of the challenge: a) Accuracy, b) L2 norm, c) ROC CA5\%. The trackers using the live ASR are represented with black bars and the trackers not using the live ASR (i.e. only the live SLU) in white. The y-ranges are adjusted to better appreciate the differences of the scores.}
\end{figure}

The featured metrics of the challenges of YARBUS on the DSTC2 and DSTC3 datasets are shown in Table~\ref{table:yarbus_dstc2} and \ref{table:yarbus_dstc3}. These are reported with the previous submissions to the challenges on the figures~\ref{fig:comparison_dstc2_test} and \ref{fig:comparison_dstc3}. The baselines (Top-hyp, Focus, HWU and HWU original) were run on the same SLUs than Yarbus and the complete set of results for these trackers is reported in Appendix A. It turns out that with its few rules and the SLU of \cite{Sun2014b}, Yarbus ranks reasonably well compared to the other approaches in the DSTC2 challenge and always better than the other baselines (top-hyp, focus, HWU and HWU+) with one exception for the ROC metric compared to HWU. For the DSTC3 challenge, Yarbus is best performing almost on the three metrics by using the error-tied SLU from \citep{Zhu2014}. It usually performs better than the other baselines.

\begin{figure}
\begin{center}
\begin{tabular}{c}
\includegraphics[width=0.7\linewidth]{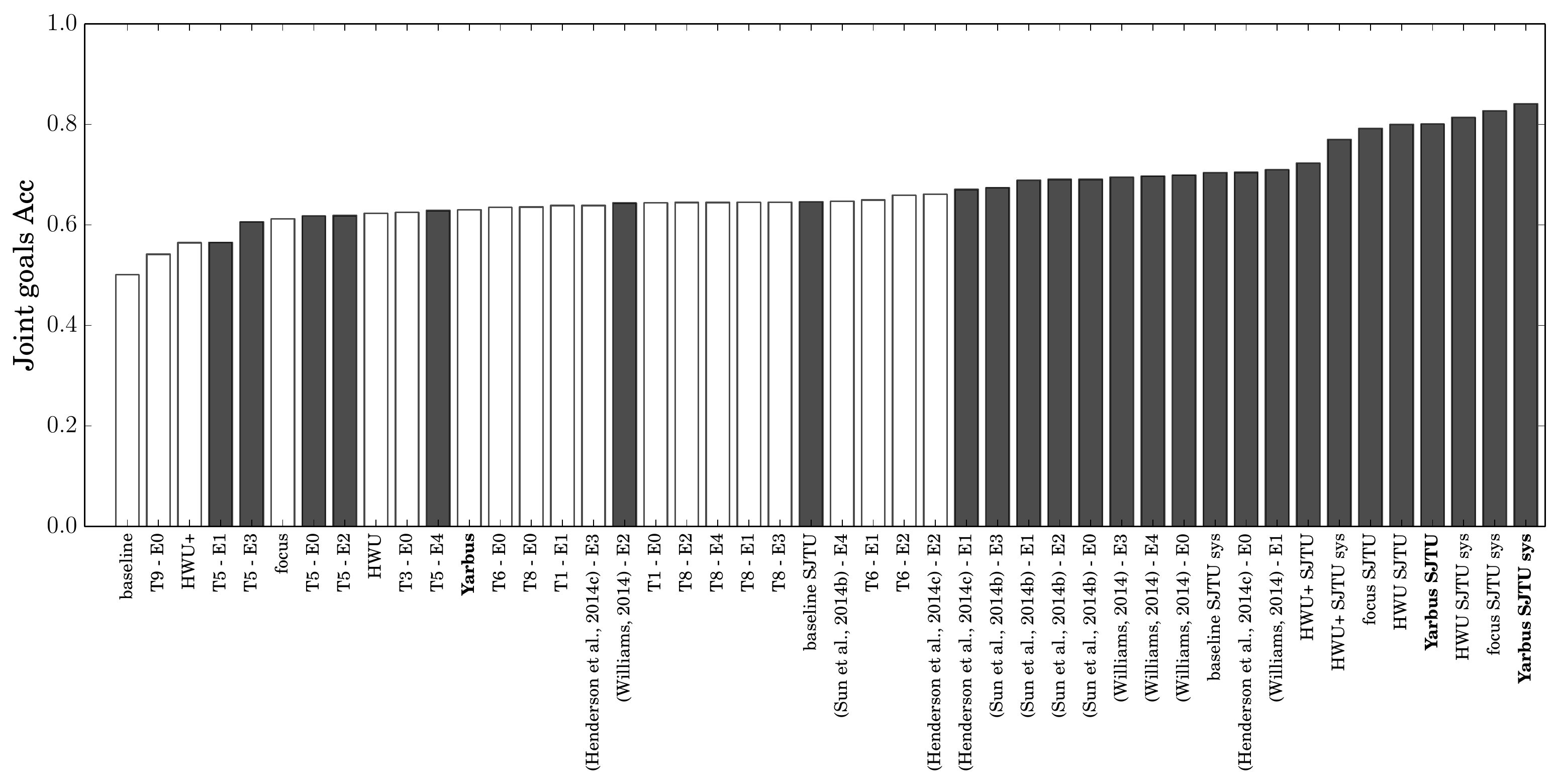}\\
a) \\
\includegraphics[width=0.7\linewidth]{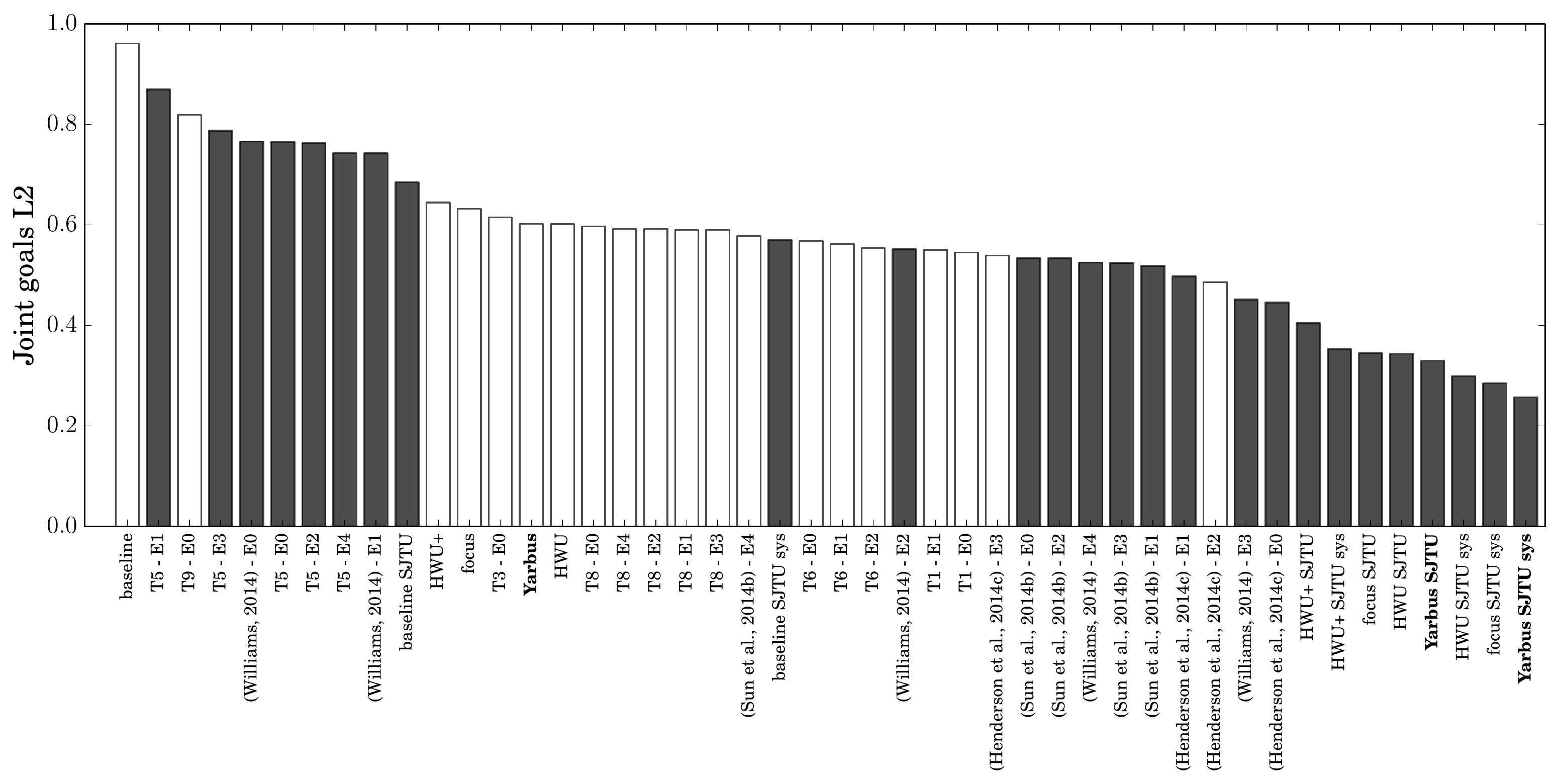}\\
b)  \\
\includegraphics[width=0.7\linewidth]{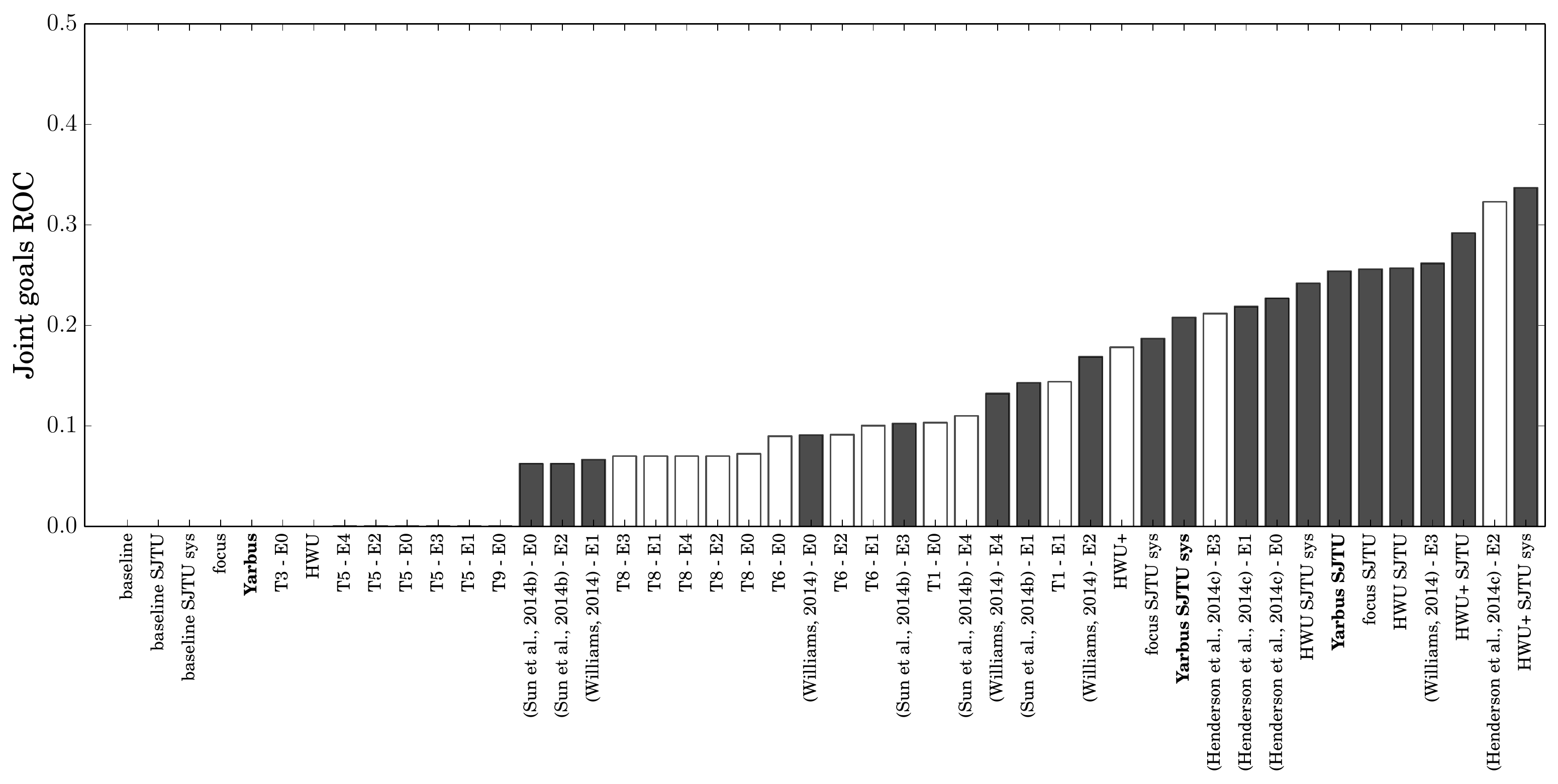}\\
c) 
\end{tabular}
\end{center}
\caption{\label{fig:comparison_dstc2_dev}Performances of the trackers on the dstc2\_dev dataset. The reported measures are the features metrics of the challenge: a) Accuracy, b) L2 norm, c) ROC CA5\%. The trackers using the live ASR are represented with black bars and the trackers not using the live ASR (i.e. only the live SLU) in white. The y-ranges are adjusted to better appreciate the differences of the scores.}
\end{figure}

Interestingly, the baselines (not only Yarbus) perform much better than the other approaches when we compare the trackers on the dstc2\_dev dataset and running the baselines on the SLU of \citep{Sun2014b}. The metrics of the trackers on the dstc2\_dev dataset are reported on Fig.~\ref{fig:comparison_dstc2_dev}.

The results of the trackers on the dstc2\_train dataset are not available except for the baselines for which the results on the three datasets of DSTC2 are provided in Appendix A. There is clear tendency for Yarbus to perform better than the other baselines on Accuracy and L2 norm but not for the ROC for which the HWU baseline is clearly better.

On the third challenge (fig.~\ref{fig:comparison_dstc3}), the difference in the results of the various trackers is much less clear than in the second challenge, at least in terms of accuracy. Yarbus is slightly better than the other baselines. The best ranking approaches are the recurrent neural network approach of \citep{Henderson2014d} and the polynomial belief tracker of~\citep{Zhu2014}.

\subsection{The size of the tracker}

\begin{center}
\begin{figure}[htbp]
\begin{center}
\begin{tabular}{cc}
 \includegraphics[width=0.3\linewidth]{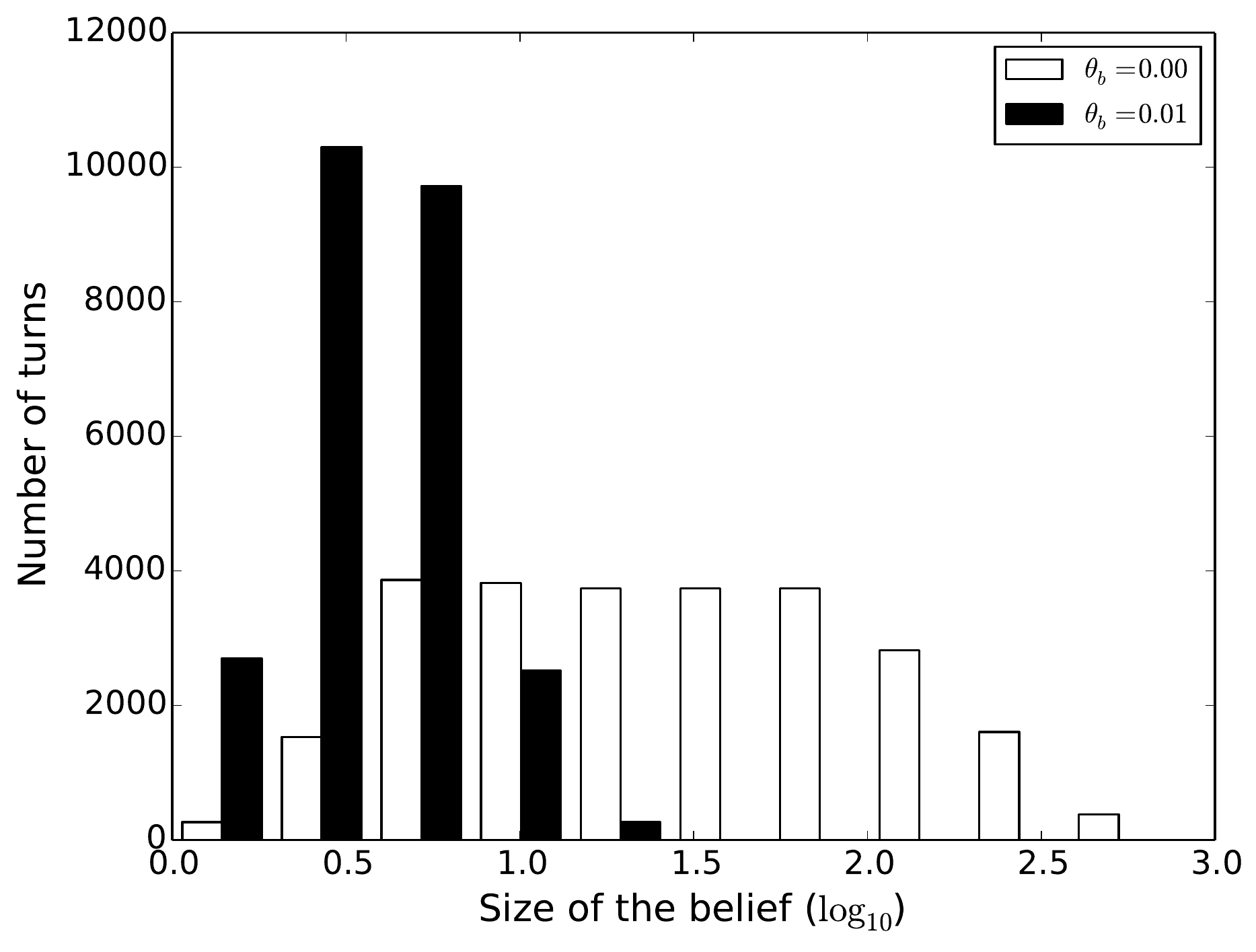}&
 \includegraphics[width=0.3\linewidth]{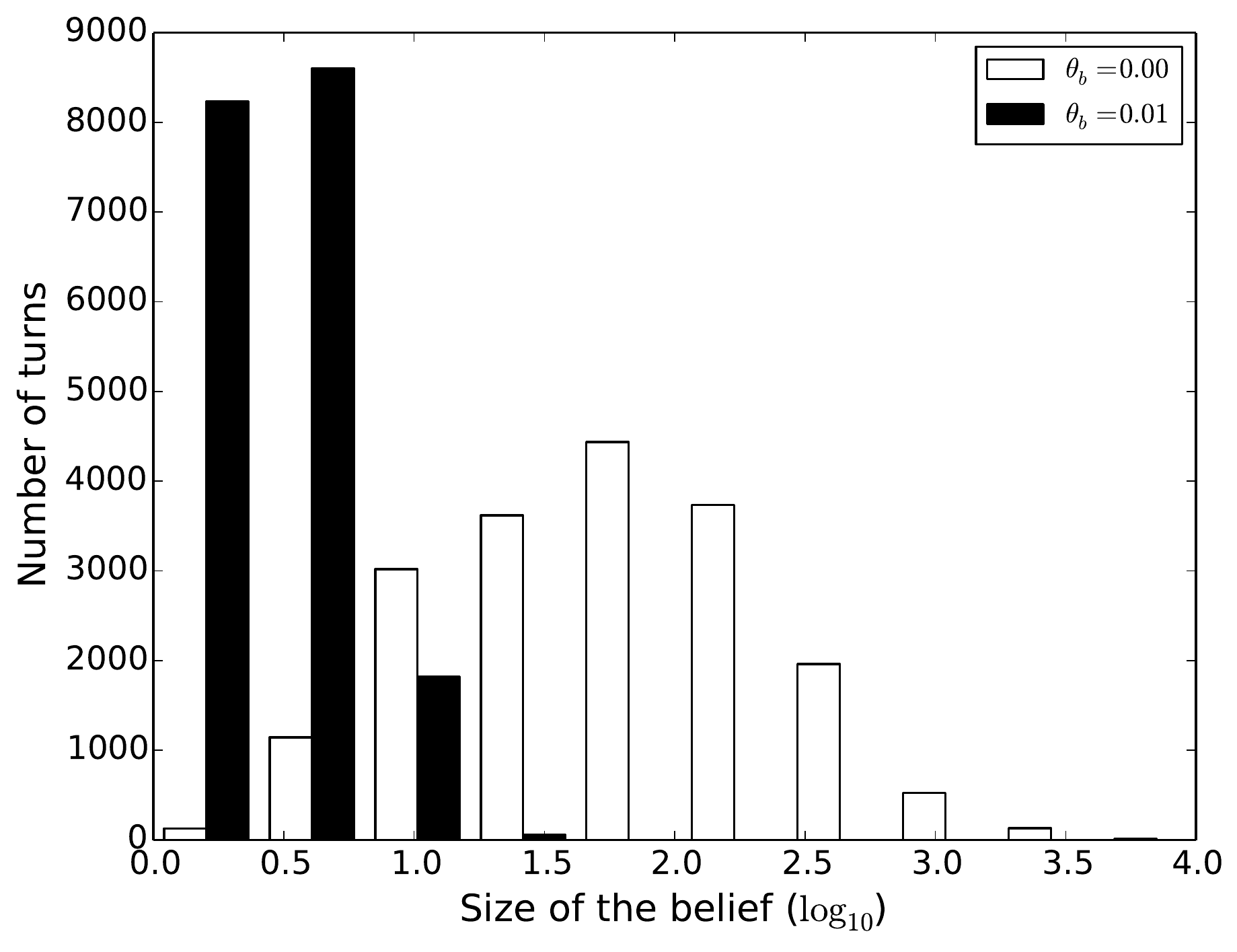} \\
a) & b)
\end{tabular}
\end{center}
\caption{\label{fig:belief_size} a) On the DSTC2 data (all the subsets included), without filtering the belief after its update, its size can grow up to $700$ joint goals. By removing the joint goals with a probability lower than $\theta_b = 10^{-2}$ and scaling the resulting belief accordingly, its size gets no more than around $30$. The results are measured on the SJTU sys SLU. b) On the dstc3\_test dataset, without filtering, the belief can have up to $10000$ entries. Filtering the belief with $\theta_b=1e-2$ significantly decreases the number of elements (no more than $30$). The experiment has been run on the SJTU err-tied SLU.}
\end{figure}
\end{center}

As noted in introduction the number of possible joint goals is much larger in the third challenge than in the second. Indeed, in this case, there are around $10^9$ possible joint goals. Therefore, when estimating the belief in the joint space, it might be that the representation becomes critically large. Such a situation is much less critical when the belief is defined in the marginal space, as in \citep{ZhuoranLemon2013} where the space to be represented is the sum and not the product of the number of values of each slot. In Yarbus, after being updated, the belief is pruned by removing all the joint goals having a probability lower than a given threshold $\theta_b$ and scaling after-while the remaining probabilities so that they sum to one. In case all the joint goals have a probability lower than $\theta_b$, the pruning is not applied as it would result in removing all the elements of the belief. If the pruning is not applied, the size of the belief might be especially large when the SLU is producing a lot of hypothesis. For example, if we measure the size of the belief on the DSTC2 challenge with the SJTU+sys SLU, the belief can contain up to $700$ entries (fig.\ref{fig:belief_size}a) and to more than $10000$ for the DSTC3 challenge with the SJTU err-tied SLU (fig.\ref{fig:belief_size}b). If the pruning is applied with $\theta_b=10^{-2}$, the number of entries in the belief does not exceed around $30$ while still keeping pretty much the same performances than the un-pruned belief. For the DSTC2 with the SJTU+sys SLU, the performances of the un-pruned belief are (Acc:0.759, L2:0.359, ROC:0.329) and the performances of the pruned belief with $\theta_b=10^{-2}$ are (Acc:0.759, L2:0.361, ROC:0.320). For the DSTC3 with the SJTU err-tied SLU, the performances of the un-pruned belief\footnote{The reported performances are actually for $\theta_b=10^{-10}$ because setting $\theta_b=0$ produced a tracker output much too large to be evaluated by the scoring scripts of the challenge.} are (Acc:0.597, L2:0.615, ROC:0.239) and the performances of the pruned belief with $\theta_b=10^{-2}$ are (Acc:0.597, L2:0.624, ROC:0.226).

%% DSTC2 + SJTU +sys
%%                         Acc      L2     ROC
%% 0.0 Yarbus             0.759  0.359   0.329     
%% 0.01 Yarbus            0.759  0.361   0.320

%% DSTC3 + SJTU err-tied
%%                         Acc      L2     ROC
%% 0.0 Yarbus      0.5967057  0.6150739    0.2392335 
%% 0.01 Yarbus            0.597	0.624	0.226

\subsection{Varying the set of rules}
\label{subsec:varying_rules}

The motivation behind the rules in Yarbus is to use a reasonably small number of rules which can hopefully extract most of the information from the machine and user acts. This choice is not driven by any dataset \emph{per se} in the sense that it might be that a smaller set of rules, which might not capture all the information from the defined acts, still performs reasonably well. That point can be checked by simply enabling or disabling rules and checking the metrics of the resulting modified Yarbus. The experimental setup is the following. Let us attribute a rule number to the five rules presented in section~\ref{sec:extracting_information} as~:
\begin{itemize}
\item Rule 0~: the $\Act{inform}$ rule in equation~(\ref{eq:pos_mus})
\item Rule 1~: the $\Act{expl-conf}$ rule in equation~(\ref{eq:pos_mus})
\item Rule 2~: the $\Act{impl-conf}$ rule in equation~(\ref{eq:pos_mus})
\item Rule 3~: the $\Act{negate}$ rule in equation~(\ref{eq:neg_mus})
\item Rule 4~: the $\Act{deny}$ rule in equation~(\ref{eq:neg_mus})
\end{itemize}

\begin{figure}[htbp]
\begin{center}
\begin{tabular}{cc}
\includegraphics[width=0.4\linewidth]{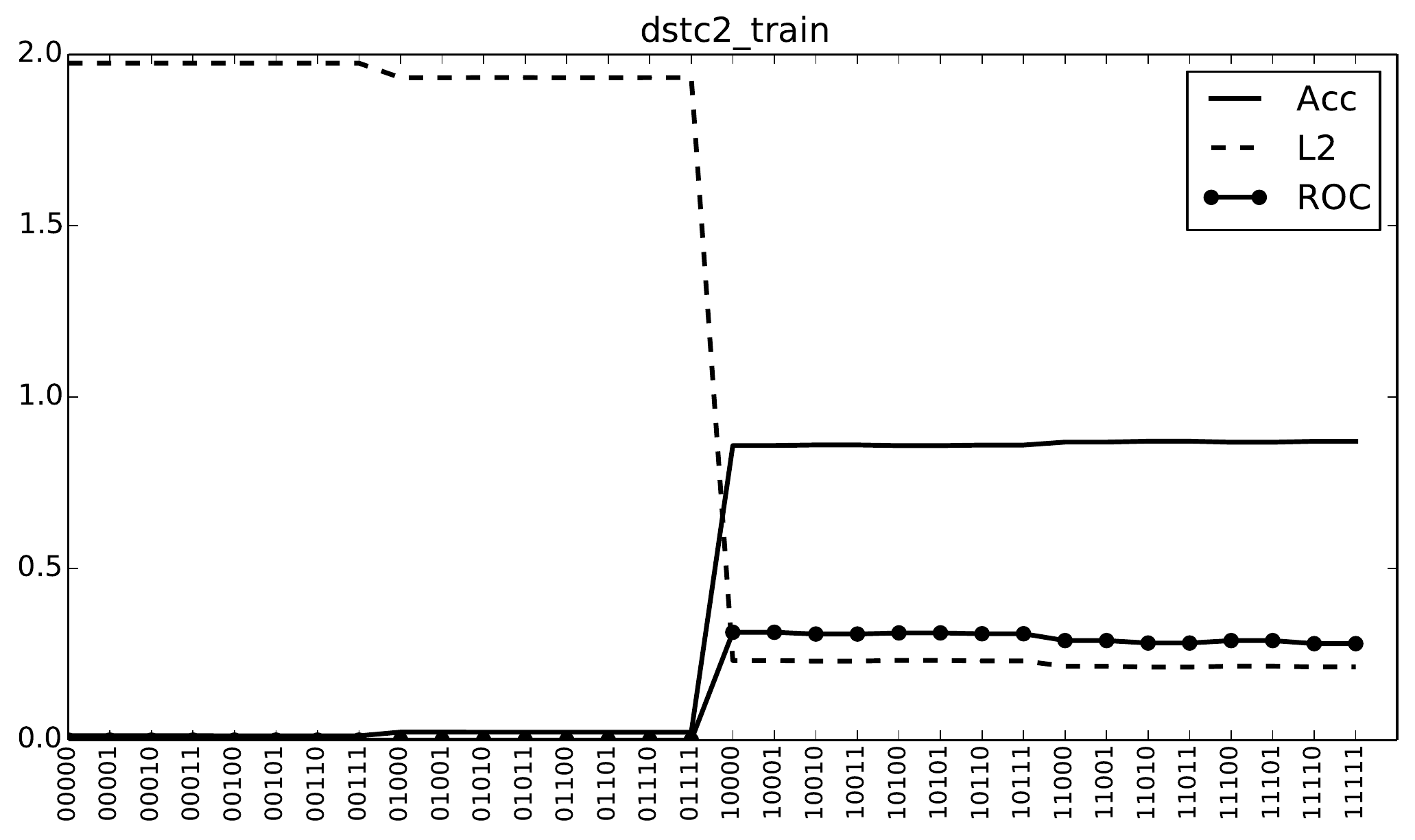}&
\includegraphics[width=0.4\linewidth]{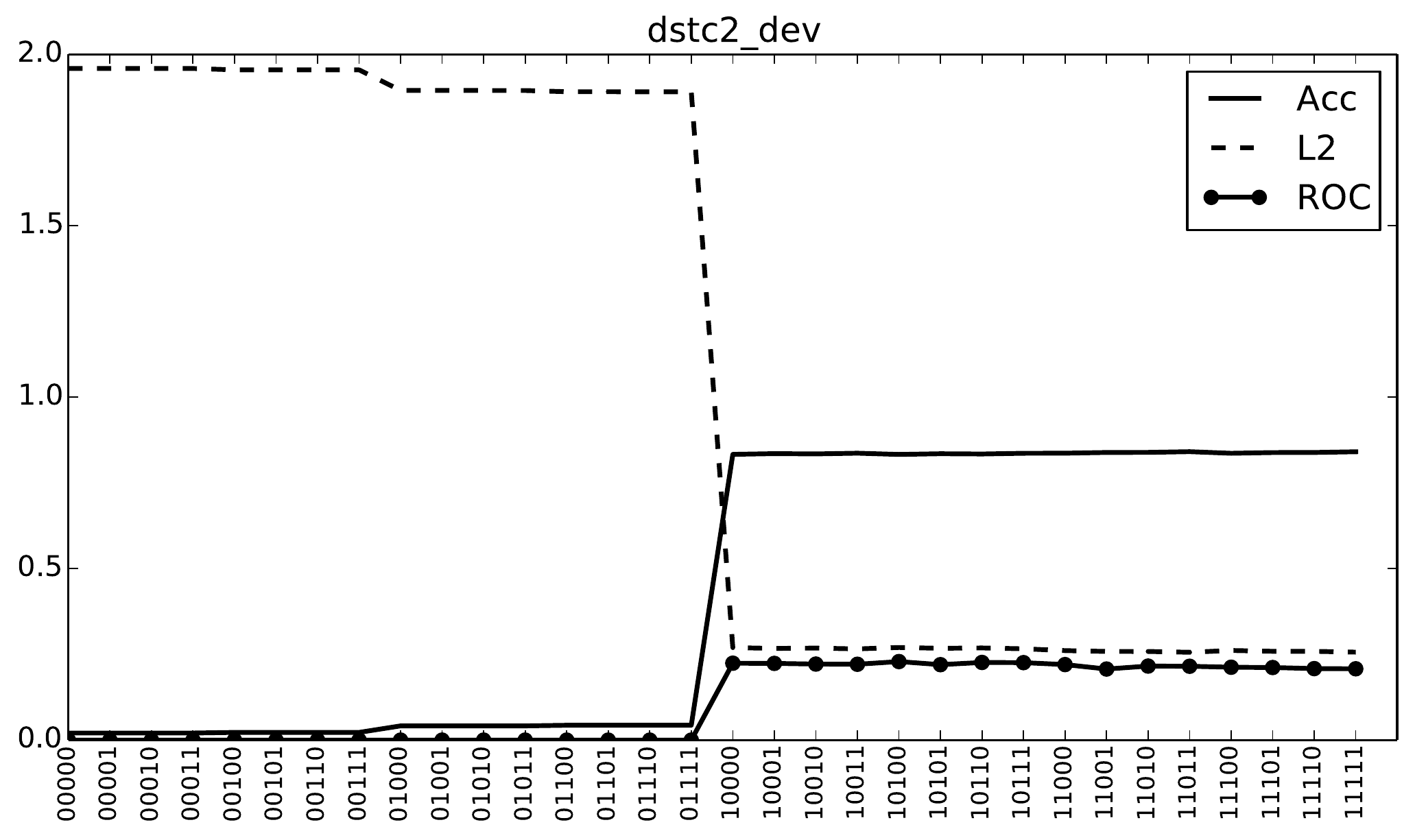}\\
a) & b)\\
\includegraphics[width=0.4\linewidth]{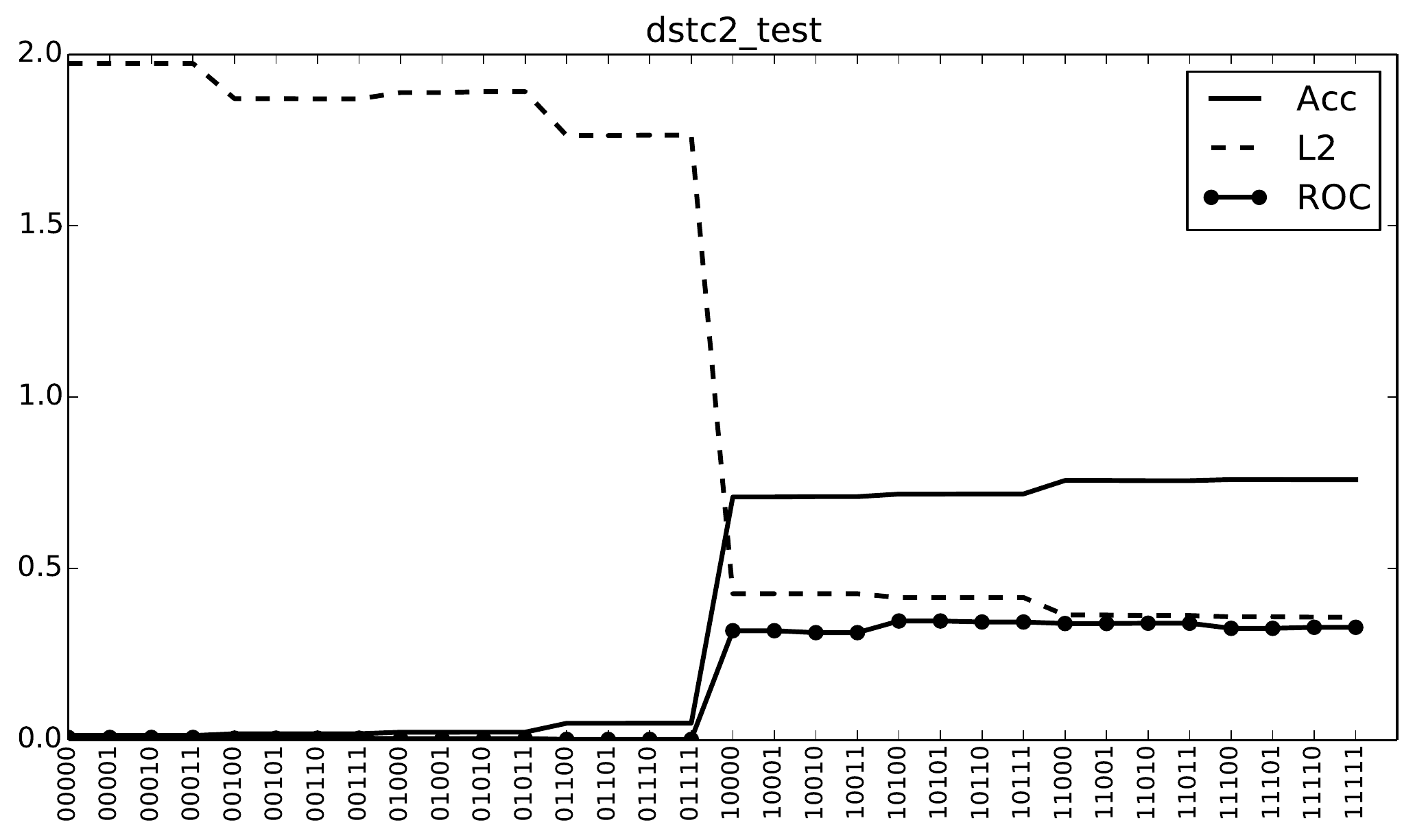}&
\includegraphics[width=0.4\linewidth]{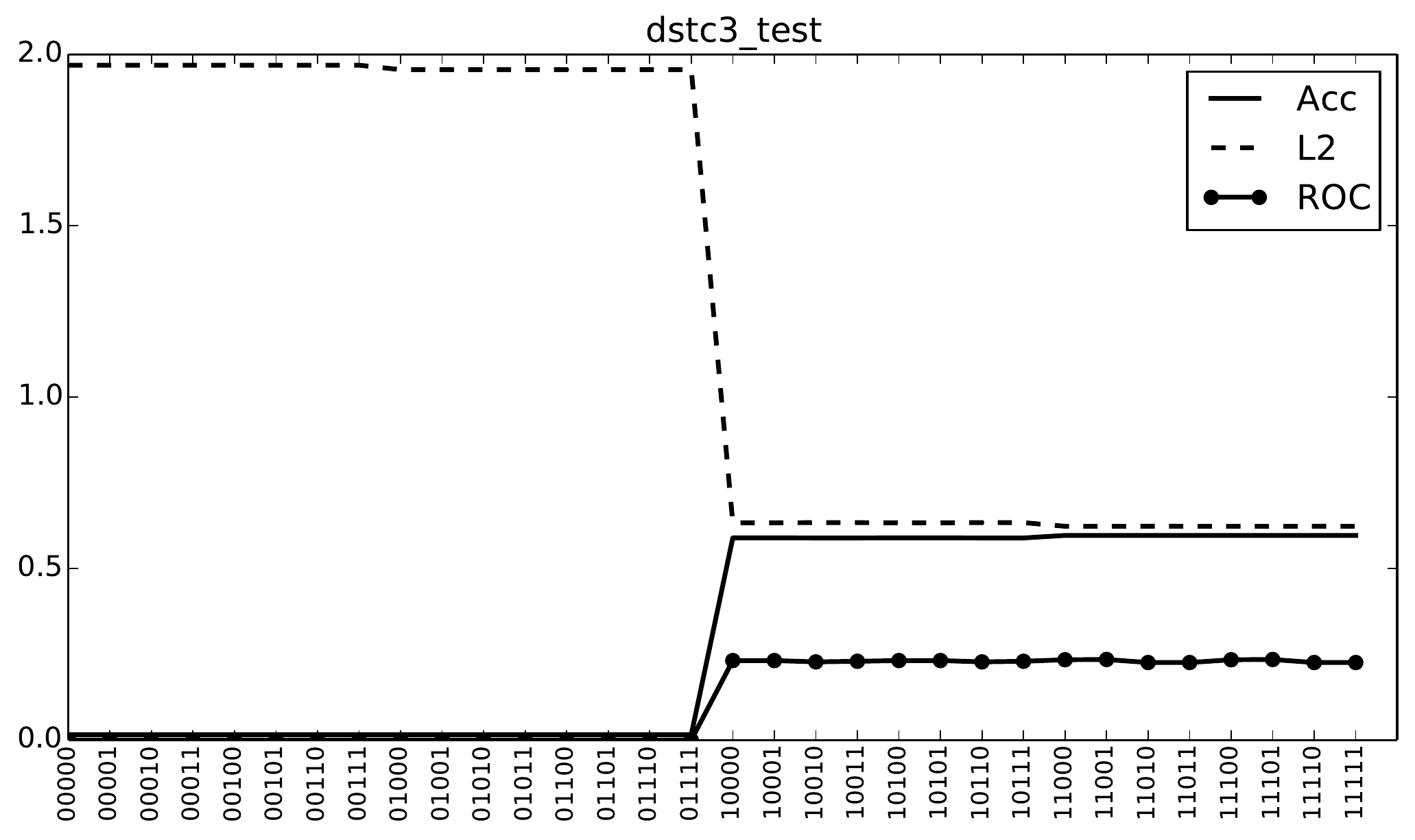}\\
c) & d)
\end{tabular}
\end{center}
\caption{\label{fig:metrics_various_sets}Metrics of YARBUS with different rule sets on a) dstc2\_train, b) dstc2\_dev, c) dstc2\_test and d) dstc3\_test.}
\end{figure}

We can now define variations of Yarbus denoted Yarbus-$r_0r_1r_2r_3r_4$ where the sequence $r_0r_1r_2r_3r_4$ identifies which rules are enabled or disabled. The tracker considered so far is therefore denoted Yarbus-$11111$. Given the $5$ rules defined above there are $32$ possibile combinations which can all be tested on the challenge datasets. In the experiment, we make use only of the SJTU+sys SLU for the DSTC2 challenge and SJTU+err-tied SLU for the DSTC3 challenge. The full set of results are given in Appendix B (Table~\ref{table:metrics_varying_rules_dstc2} and \ref{table:metrics_varying_rules_dstc3}). The metrics of the various trackers on the different datasets are plotted on Fig.~\ref{fig:metrics_various_sets}. It turns out that a big step in the metrics is obtained when enabling the Rule 0, i.e. the inform rule which is not much a surprise. However it might be noted that the performances do not grow much by adding additional rules. There is one exception for the dstc2\_test dataset for which the inclusion of the second rule leads to an increase in $5 \%$ in accuracy. The conclusion is clearly that even if the additional rules make sense in the information they capture, they do not lead to significant improvements on the metrics and the performances can be obtained by making use of only two rules : the inform and expl-conf rules of equation~(\ref{eq:pos_mus}).

\section{Discussion}
\label{sec:discussion}

In this paper, we presented a rule-based belief tracker. The tracker, which does not require any learning, performs favourably on the dialog state tracking challenges in comparison with other approaches. It should be noted that there is a significant increase in performances by switching from the live SLU of the challenges to the SLU of \citep{Sun2014b,Zhu2014}. Yarbus is a very simple tracker. We tried actually to add rules that appeared at first glance to capture more information or to capture the information in a more coherent way (for example by considering alternatives in the way the reference of $\SlotFormat{this}$ is solved) but these attempts resulted in degraded performances. Yarbus is in no way a very tricky belief tracker. Most of the \emph{expertise} comes from the design of the rules extracting information from the utterances but otherwise the update of the belief is based on simple Bayesian rules. In light of the results of section \ref{subsec:varying_rules}, it turns out that Yarbus could be even simpler by considering only two out of the five rules. However, the point of the paper was clearly not to devise a new belief tracker but the fact that Yarbus uses rules involving the dialog acts is actually quite informative. First, despite its simplicity, if one compares the performances of Yarbus with the best ranking tracker proposed by \citep{Williams2014}, there are dialogs on which the first performs better than the second and dialogs on which the second performs better than the first. In that respect, Yarbus might still be a good candidate for ensemble learning \citep{Henderson2014a}. Second, using such a simple rule based tracker informs us on the real performances of more elaborate machine learning based techniques such as recurrent neural networks \citep{Henderson2014b} or deep neural networks \citep{Henderson2013a}. These latter techniques are rather blind to the data being processed. Even if these approaches performs well at first sight, the performances of Yarbus allow to better appreciate what part of the information is really extracted from the data. Last, one natural conclusion from the results of this paper is that there is still work to be done in order to get real breakthroughs in slot filling tasks. Since a simple rule based system performs very well (more than $75 \%$ of accuracy) on the second challenge is raising the question of making use of this dataset for evaluating belief trackers. On the third challenge, the conclusion is less straightforward. However, it is clear from the datasets of DSTC2 and DSTC3 that the biggest improvements were achieved thanks to the SLU and this suggests to shift the focus on this element of the dialog loop.

%% \begin{itemize}
%% \item a rule based system focuses on the very nature of the data
%% \item on a mis plus de règles mais ça merdait
%% \item ces règles là ça marche bien mais la dernière section nous enseigne que les règles 1 et 2 suffiraient à expliquer la plus grosse partie des scores
%% \item d'ou la question de savoir ce que ça veut dire de réussir ce challenge, au final, les trackers arrivent à faire à peu près ce que fait une règle, d'ou d'ailleurs l'avantage d'utiliser ce rule based, c'est au moins ce qu'il nous apprend 
%% \end{itemize}

\paragraph{Acknowledgments:} This work has been supported by l'Agence Nationale pour la Recherche under projet reference ANR-12-CORD-0021 (MaRDi).

%%bibliographystyle is included in jmlr2e-jddr

%\bibliography{biblio.bib}

\begin{thebibliography}{15}
\providecommand{\natexlab}[1]{#1}
\providecommand{\url}[1]{\texttt{#1}}
\expandafter\ifx\csname urlstyle\endcsname\relax
  \providecommand{\doi}[1]{doi: #1}\else
  \providecommand{\doi}{doi: \begingroup \urlstyle{rm}\Url}\fi

\bibitem[Black et~al.(2011)Black, Burger, Conkie, Hastie, Keizer, Lemon,
  Merigaud, Parent, Schubiner, Thomson, Williams, Yu, Young, and
  Eskenazi]{Black2010}
Alan~W Black, Susanne Burger, Alistair Conkie, Helen Hastie, Simon Keizer,
  Oliver Lemon, Nicolas Merigaud, Gabriel Parent, Gabriel Schubiner, Blaise
  Thomson, Jason~D. Williams, Kai Yu, Steve Young, and Maxine Eskenazi.
\newblock Spoken dialog challenge 2010: Comparison of live and control test
  results.
\newblock In \emph{Proc SIGDIAL, Portland, Oregon, USA}, 2011.
\newblock URL
  \url{http://research.microsoft.com/apps/pubs/default.aspx?id=160915}.

\bibitem[Fix and Frezza-Buet(2015)]{yarbussrc}
Jeremy Fix and Herv{\'e} Frezza-Buet.
\newblock Yarbus source code, 2015.
\newblock URL \url{https://github.com/jeremyfix/dstc/}.

\bibitem[Goddeau et~al.(1996)Goddeau, Meng, Polifroni, Seneff, and
  Busayapongchai]{Goddeau1996}
D.~Goddeau, H.~Meng, J.~Polifroni, S.~Seneff, and S.~Busayapongchai.
\newblock A form-based dialogue manager for spoken language applications.
\newblock In \emph{Spoken Language, 1996. ICSLP 96. Proceedings., Fourth
  International Conference on}, volume~2, pages 701--704 vol.2, Oct 1996.
\newblock \doi{10.1109/ICSLP.1996.607458}.
\newblock URL \url{http://dx.doi.org/10.1109/ICSLP.1996.607458}.

\bibitem[Henderson et~al.(2013)Henderson, Thomson, and Young]{Henderson2013a}
M.~Henderson, B.~Thomson, and S.~J. Young.
\newblock {Deep Neural Network Approach for the Dialog State Tracking
  Challenge}.
\newblock In \emph{Proceedings of SIGdial}, 2013.

\bibitem[Henderson et~al.(2014{\natexlab{a}})Henderson, Thomson, and
  Young]{Henderson2014d}
M.~Henderson, B.~Thomson, and S.~J. Young.
\newblock {Robust Dialog State Tracking Using Delexicalised Recurrent Neural
  Networks and Unsupervised Adaptation}.
\newblock In \emph{Proceedings of IEEE Spoken Language Technology},
  2014{\natexlab{a}}.
\newblock URL \url{http://mi.eng.cam.ac.uk/~sjy/papers/htyo14.pdf}.

\bibitem[Henderson et~al.(2013--2014)Henderson, Thomson, and
  Williams]{dstchandbook}
Matthew Henderson, Blaise Thomson, and Jason Williams.
\newblock Dialog state tracking challenge 2 \& 3, 2013--2014.
\newblock URL \url{http://camdial.org/~mh521/dstc/}.

\bibitem[Henderson et~al.(2014{\natexlab{b}})Henderson, Thomson, and
  Williams]{Henderson2014a}
Matthew Henderson, Blaise Thomson, and Jason~D Williams.
\newblock The second dialog state tracking challenge.
\newblock In \emph{Proceedings of the 15th Annual Meeting of the Special
  Interest Group on Discourse and Dialogue (SIGDIAL)}, pages 263--272,
  Philadelphia, PA, U.S.A., June 2014{\natexlab{b}}. Association for
  Computational Linguistics.
\newblock URL \url{http://www.aclweb.org/anthology/W14-4337}.

\bibitem[Henderson et~al.(2014{\natexlab{c}})Henderson, Thomson, and
  Young]{Henderson2014b}
Matthew Henderson, Blaise Thomson, and Steve Young.
\newblock Word-based dialog state tracking with recurrent neural networks.
\newblock In \emph{Proceedings of the 15th Annual Meeting of the Special
  Interest Group on Discourse and Dialogue (SIGDIAL)}, pages 292--299,
  Philadelphia, PA, U.S.A., June 2014{\natexlab{c}}. Association for
  Computational Linguistics.
\newblock URL \url{http://www.aclweb.org/anthology/W14-4340}.

\bibitem[Rieser and Lemon(2011)]{Rieser2011}
Verena Rieser and Oliver Lemon.
\newblock \emph{Reinforcement Learning for Adaptive Dialogue Systems: A
  Data-driven Methodology for Dialogue Management and Natural Language
  Generation}.
\newblock Springer-Verlag, 2011.

\bibitem[Sun et~al.(2014{\natexlab{a}})Sun, Chen, Zhu, and Yu]{Sun2014a}
Kai Sun, Lu~Chen, Su~Zhu, and Kai Yu.
\newblock A generalized rule based tracker for dialogue state tracking.
\newblock In \emph{Proceedings 2014 IEEE Spoken Language Technology Workshop},
  South Lake Tahoe, USA, December 2014{\natexlab{a}}.
\newblock URL \url{http://www.aiexp.info/files/slt_1056.pdf}.

\bibitem[Sun et~al.(2014{\natexlab{b}})Sun, Chen, Zhu, and Yu]{Sun2014b}
Kai Sun, Lu~Chen, Su~Zhu, and Kai Yu.
\newblock The sjtu system for dialog state tracking challenge 2.
\newblock In \emph{Proceedings of the 15th Annual Meeting of the Special
  Interest Group on Discourse and Dialogue (SIGDIAL)}, pages 318--326,
  Philadelphia, PA, U.S.A., June 2014{\natexlab{b}}. Association for
  Computational Linguistics.
\newblock URL
  \url{http://www.sigdial.org/workshops/conference15/proceedings/pdf/W14-4343.pdf}.

\bibitem[Wang and Lemon(2013)]{ZhuoranLemon2013}
Zhuoran Wang and Oliver Lemon.
\newblock A simple and generic belief tracking mechanism for the dialog state
  tracking challenge: On the believability of observed information.
\newblock In \emph{Proceedings of the SIGDIAL 2013 Conference}, pages 423--432,
  Metz, France, August 2013. Association for Computational Linguistics.
\newblock URL \url{http://www.aclweb.org/anthology/W/W13/W13-4067}.

\bibitem[Williams et~al.(2013)Williams, Raux, Ramachandran, and
  Black]{Williams2013}
Jason Williams, Antoine Raux, Deepak Ramachandran, and Alan Black.
\newblock The dialog state tracking challenge.
\newblock In \emph{Proceedings of the SIGDIAL 2013 Conference}, pages 404--413,
  Metz, France, August 2013. Association for Computational Linguistics.
\newblock URL \url{http://www.aclweb.org/anthology/W/W13/W13-4065}.

\bibitem[Williams(2014)]{Williams2014}
Jason~D Williams.
\newblock Web-style ranking and slu combination for dialog state tracking.
\newblock In \emph{Proceedings of the 15th Annual Meeting of the Special
  Interest Group on Discourse and Dialogue (SIGDIAL)}, pages 282--291,
  Philadelphia, PA, U.S.A., June 2014. Association for Computational
  Linguistics.
\newblock URL \url{http://www.aclweb.org/anthology/W14-4339}.

\bibitem[Zhu et~al.(2014)Zhu, Chen, Sun, Zheng, and Yu]{Zhu2014}
Su~Zhu, Lu~Chen, Kai Sun, Da~Zheng, and Kai Yu.
\newblock Semantic parser enhancement for dialogue domain extension with little
  data.
\newblock In \emph{Proceedings 2014 IEEE Spoken Language Technology Workshop},
  South Lake Tahoe, USA, 2014.
\newblock URL \url{http://www.aiexp.info/files/slt_1062.pdf}.

\end{thebibliography}

\newpage
\pagebreak

\section*{Appendix A : Scores of the different baselines with the various SLU}

\begin{table}[htbp]
\begin{center}
\begin{footnotesize}
\begin{tabular}{|l||c|c|c|c|c|c|c|c|c|}
\hline
          & \multicolumn{9}{c|}{Live SLU}\\ \cline{2-10}
        &\multicolumn{3}{c|}{dstc2\_train} & \multicolumn{3}{c|}{dstc2\_dev} & \multicolumn{3}{c|}{dstc2\_test} \\ \cline{2-10}
          & Acc. & L2 & ROC ca5      & Acc. & L2 & ROC ca5     &  Acc. & L2 & ROC ca5 \\
\hline
top-hyp   & 0.582& 0.810& 0& 0.501& 0.961& 0& 0.619& 0.738& 0\\
focus     & 0.715& 0.471& 0& 0.612& 0.632& 0& 0.719& 0.464& 0\\
HWU       & 0.732& 0.451& 0& 0.623& 0.601& 0 & 0.711& 0.466& 0 \\
HWU+      & 0.646& 0.518& 0.185& 0.564& 0.645& 0.178& 0.666& 0.498& 0.210 \\
Yarbus    & 0.719& 0.464& 0&0.630 &0.602 & 0& 0.725 & 0.440& 0  \\
\hline\hline
         & \multicolumn{9}{c|}{SJTU 1best SLU}\\ \cline{2-10}
        &\multicolumn{3}{c|}{dstc2\_train} & \multicolumn{3}{c|}{dstc2\_dev} & \multicolumn{3}{c|}{dstc2\_test} \\ \cline{2-10}
          & Acc. & L2 & ROC ca5      & Acc. & L2 & ROC ca5     &  Acc. & L2 & ROC ca5 \\
\hline
top-hyp  & 0.656& 0.669& 0.032& 0.646& 0.685& 0& 0.600& 0.771& 0.23\\
focus    & 0.831& 0.278& 0.220& 0.792& 0.345& 0.256& 0.740& 0.405& 0.254\\
HWU      & 0.841& 0.266& 0.206& 0.800& 0.344& 0.257& 0.737& 0.413& 0.287 \\
HWU+     & 0.755& 0.337& 0.318& 0.723& 0.405& 0.292& 0.703& 0.449& 0.252\\
Yarbus   & 0.835& 0.265& 0.232& 0.801& 0.330& 0.254& 0.752 & 0.392& 0.271  \\
\hline\hline
          & \multicolumn{9}{c|}{SJTU 1best+sys SLU}\\ \cline{2-10}
        &\multicolumn{3}{c|}{dstc2\_train} & \multicolumn{3}{c|}{dstc2\_dev} & \multicolumn{3}{c|}{dstc2\_test} \\ \cline{2-10}
          & Acc. & L2 & ROC ca5      & Acc. & L2 & ROC ca5     &  Acc. & L2 & ROC ca5 \\
\hline
top-hyp & 0.722& 0.512& 0.038& 0.704& 0.570& 0 & 0.622& 0.728& 0.020\\
focus   & 0.862& 0.231& 0.273& 0.827& 0.285& 0.187& 0.745 & 0.371& 0.320\\
HWU     & 0.859& 0.231& 0.299& 0.814& 0.299& 0.242& 0.730& 0.396& 0.359\\
HWU+    & 0.803& 0.300& 0.380& 0.770& 0.353& 0.337& 0.716 & 0.436& 0.322\\
Yarbus  & 0.871& 0.213& 0.281& 0.841& 0.257& 0.208&0.759 & 0.359&  0.329\\
\hline
\end{tabular}
\end{footnotesize}
\end{center}
\caption{\label{table:soa_dstc2_joint}Results for the joint goals on the DSCT2 challenge. HWU denotes the Heriot-Watt tracker\citep{ZhuoranLemon2013}, HWU+ is with the original flag enabled.}
\end{table}

\begin{table}[htbp]
\begin{center}
\begin{footnotesize}
\begin{tabular}{|l||c|c|c||c|c|c|}
\hline
        & \multicolumn{3}{c||}{Live SLU} & \multicolumn{3}{c|}{SJTU asr-tied SLU}\\ 
\cline{2-7}
        & Acc. & L2 & ROC ca5      & Acc. & L2 & ROC ca5  \\
\hline
top-hyp                 & 0.555& 0.860& 0& 0.591& 0.778& 0.116\\
focus                    & 0.556& 0.750& 0& 0.589 & 0.632& 0.274\\
HWU                      & 0.575& 0.744& 0& - & - & - \\
HWU+                     & 0.567& 0.691& 0& - & - & - \\
Yarbus $\theta_b=10^{-2}$ & 0.582& 0.702& 0&0.597 &0.624 &0.226\\
\hline\hline
        & \multicolumn{3}{c||}{SJTU errgen SLU} & \multicolumn{3}{c|}{SJTU errgen+rescore SLU}\\ 
\cline{2-7}
        & Acc. & L2 & ROC ca5      & Acc. & L2 & ROC ca5  \\
\hline
top-hyp                 & 0.588& 0.779& 0.114& 0.587& 0.780& 0.122\\
focus                    & 0.587& 0.623& 0.218& 0.579& 0.613&  0.225\\
HWU                      & - & - & - & - & - & -\\
HWU+                     & - & - & - & - & - & -\\
Yarbus $\theta_b=10^{-2}$ & 0.594& 0.624& 0.150& 0.595& 0.607&0.151\\
\hline
\end{tabular}
\end{footnotesize}
\end{center}
\caption{\label{table:soa_dstc3_joint}Results for the joint goals on the DSCT3 challenge. For HWU and HWU+, using the SJTU SLUs leads to a tracker output too large to hold in memory and the results are therefore not available on these datasets.}
\end{table}

%% \begin{table}[htbp]
%% \begin{center}
%% \begin{tabular}{|r||c|c|c|}
%% \hline 
%%         & \multicolumn{3}{c|}{dstc3\_test - Joint goals}\\
%% \hline
%%         & Acc. & L2 & ROC ca5 \\
%% \hline
%% $DNN$ \citep{Sun2014a} &0.583?? &0.583??? & \\
%% $f^\star$   &0.606?? & 0.561??& \\
%% \hline
%% RNN \citep{Henderson2014d} T3E0& 0.646&  0.538& ??\\
%% \hline
%% \end{tabular}
%% \end{center}
%% \caption{\label{table:soa_dstc3_joint_all}Results for the joint goals on the DSCT3 challenge.}
%% \end{table}

\pagebreak
\newpage

\section*{Appendix B: Metrics of YARBUS with various rule sets}

\begin{table}[hbp]
\begin{tabular}{cc}
\begin{tabular}{|c|c|c|c|}
\hline
\textbf{Rule set} & \textbf{Accuracy} & \textbf{L2}& \textbf{ROC ca5} \\ 
\hline 
00000&0.0128014&1.9743972&0.0000000\\ 
\hline 
00001&0.0128014&1.9743972&0.0000000\\ 
\hline 
00010&0.0128014&1.9743972&0.0000000\\ 
\hline 
00011&0.0128014&1.9743972&0.0000000\\ 
\hline 
00100&0.0125384&1.9742266&0.0000000\\ 
\hline 
00101&0.0125384&1.9742266&0.0000000\\ 
\hline 
00110&0.0125384&1.9742266&0.0000000\\ 
\hline 
00111&0.0125384&1.9742266&0.0000000\\ 
\hline 
01000&0.0233231&1.9318885&0.0000000\\ 
\hline 
01001&0.0233231&1.9318885&0.0000000\\ 
\hline 
01010&0.0230601&1.9323025&0.0000000\\ 
\hline 
01011&0.0230601&1.9323025&0.0000000\\ 
\hline 
01100&0.0230601&1.9317179&0.0000000\\ 
\hline 
01101&0.0230601&1.9317179&0.0000000\\ 
\hline 
01110&0.0227970&1.9321319&0.0000000\\ 
\hline 
01111&0.0227970&1.9321319&0.0000000\\ 
\hline 
10000&0.8590969&0.2320063&0.3144519\\ 
\hline 
10001&0.8590969&0.2320063&0.3144519\\ 
\hline 
10010&0.8608505&0.2303032&0.3093298\\ 
\hline 
10011&0.8608505&0.2303032&0.3093298\\ 
\hline 
10100&0.8587462&0.2324565&0.3126404\\ 
\hline 
10101&0.8587462&0.2324565&0.3126404\\ 
\hline 
10110&0.8604998&0.2307534&0.3101691\\ 
\hline 
10111&0.8604998&0.2307534&0.3101691\\ 
\hline 
11000&0.8690925&0.2153537&0.2902542\\ 
\hline 
11001&0.8690925&0.2153537&0.2902542\\ 
\hline 
11010&0.8718106&0.2127873&0.2831137\\ 
\hline 
11011&0.8718106&0.2127873&0.2831137\\ 
\hline 
11100&0.8687418&0.2158038&0.2902705\\ 
\hline 
11101&0.8687418&0.2158038&0.2902705\\ 
\hline 
11110&0.8714599&0.2132374&0.2813160\\ 
\hline 
11111&0.8714599&0.2132374&0.2813160\\ 
\hline 
\end{tabular}
& 
\begin{tabular}{|c|c|c|c|}
\hline
\textbf{Rule set} & \textbf{Accuracy} & \textbf{L2}& \textbf{ROC ca5} \\ 
\hline 
00000&0.0205944&1.9588113&0.0000000\\ 
\hline 
00001&0.0205944&1.9588113&0.0000000\\ 
\hline 
00010&0.0205944&1.9588113&0.0000000\\ 
\hline 
00011&0.0205944&1.9588113&0.0000000\\ 
\hline 
00100&0.0224192&1.9549719&0.0000000\\ 
\hline 
00101&0.0224192&1.9549719&0.0000000\\ 
\hline 
00110&0.0224192&1.9549719&0.0000000\\ 
\hline 
00111&0.0224192&1.9549719&0.0000000\\ 
\hline 
01000&0.0414494&1.8949548&0.0000000\\ 
\hline 
01001&0.0414494&1.8949548&0.0000000\\ 
\hline 
01010&0.0414494&1.8946709&0.0000000\\ 
\hline 
01011&0.0414494&1.8946709&0.0000000\\ 
\hline 
01100&0.0432742&1.8911155&0.0000000\\ 
\hline 
01101&0.0432742&1.8911155&0.0000000\\ 
\hline 
01110&0.0432742&1.8908316&0.0000000\\ 
\hline 
01111&0.0432742&1.8908316&0.0000000\\ 
\hline 
10000&0.8334202&0.2697380&0.2242728\\ 
\hline 
10001&0.8355057&0.2673630&0.2237129\\ 
\hline 
10010&0.8347237&0.2683460&0.2217364\\ 
\hline 
10011&0.8368092&0.2659710&0.2211838\\ 
\hline 
10100&0.8331595&0.2701215&0.2290363\\ 
\hline 
10101&0.8352450&0.2677466&0.2200375\\ 
\hline 
10110&0.8344630&0.2687295&0.2264917\\ 
\hline 
10111&0.8365485&0.2663545&0.2259271\\ 
\hline 
11000&0.8368092&0.2609925&0.2202492\\ 
\hline 
11001&0.8388947&0.2586176&0.2072716\\ 
\hline 
11010&0.8391554&0.2585226&0.2159056\\ 
\hline 
11011&0.8412409&0.2561476&0.2153703\\ 
\hline 
11100&0.8365485&0.2613761&0.2125273\\ 
\hline 
11101&0.8386340&0.2590011&0.2116879\\ 
\hline 
11110&0.8388947&0.2589061&0.2085146\\ 
\hline 
11111&0.8409802&0.2565311&0.2079975\\ 
\hline 
\end{tabular} \\
a) & b)
\end{tabular}
\caption{\label{table:metrics_varying_rules_dstc2}Metrics of YARBUS with various rule sets on a) the dstc2\_train and b) the dstc2\_dev datasets. The meaning of the rule set number is defined in section~\ref{subsec:varying_rules}. The trackers were run on the SJTU + sys SLU.}
\end{table}

\begin{table}[h!]
\begin{tabular}{cc}
\begin{tabular}{|c|c|c|c|}
\hline
\textbf{Rule set} & \textbf{Accuracy} & \textbf{L2}& \textbf{ROC ca5} \\ 
\hline 
00000&0.0132109&1.9735783&0.0078125\\ 
\hline 
00001&0.0132109&1.9735783&0.0078125\\ 
\hline 
00010&0.0132109&1.9735783&0.0078125\\ 
\hline 
00011&0.0132109&1.9735783&0.0078125\\ 
\hline 
00100&0.0184746&1.8709733&0.0055866\\ 
\hline 
00101&0.0184746&1.8709733&0.0055866\\ 
\hline 
00110&0.0184746&1.8703227&0.0055866\\ 
\hline 
00111&0.0184746&1.8703227&0.0055866\\ 
\hline 
01000&0.0230158&1.8886356&0.0044843\\ 
\hline 
01001&0.0230158&1.8886356&0.0044843\\ 
\hline 
01010&0.0231190&1.8913796&0.0044643\\ 
\hline 
01011&0.0231190&1.8913796&0.0044643\\ 
\hline 
01100&0.0491279&1.7631903&0.0021008\\ 
\hline 
01101&0.0491279&1.7631903&0.0021008\\ 
\hline 
01110&0.0492311&1.7644893&0.0020964\\ 
\hline 
01111&0.0492311&1.7644893&0.0020964\\ 
\hline 
10000&0.7091547&0.4269933&0.3190220\\ 
\hline 
10001&0.7091547&0.4269933&0.3190220\\ 
\hline 
10010&0.7100836&0.4268610&0.3133721\\ 
\hline 
10011&0.7100836&0.4268610&0.3133721\\ 
\hline 
10100&0.7175147&0.4156691&0.3473820\\ 
\hline 
10101&0.7175147&0.4156691&0.3473820\\ 
\hline 
10110&0.7179275&0.4158861&0.3444508\\ 
\hline 
10111&0.7179275&0.4158861&0.3444508\\ 
\hline 
11000&0.7572505&0.3647205&0.3400572\\ 
\hline 
11001&0.7572505&0.3647205&0.3400572\\ 
\hline 
11010&0.7566312&0.3632813&0.3408812\\ 
\hline 
11011&0.7566312&0.3632813&0.3408812\\ 
\hline 
11100&0.7598307&0.3595799&0.3259984\\ 
\hline 
11101&0.7598307&0.3595799&0.3259984\\ 
\hline 
11110&0.7592115&0.3585286&0.3289831\\ 
\hline 
11111&0.7592115&0.3585286&0.3289831\\ 
\hline 
\end{tabular}&
\begin{tabular}{|c|c|c|c|}
\hline 
\textbf{Rule set} & \textbf{Accuracy} & \textbf{L2}& \textbf{ROC ca5} \\ 
\hline 
00000&0.0157356&1.9685289&0.0000000\\ 
\hline 
00001&0.0157356&1.9685289&0.0000000\\ 
\hline 
00010&0.0157356&1.9685289&0.0000000\\ 
\hline 
00011&0.0157356&1.9685289&0.0000000\\ 
\hline 
00100&0.0157356&1.9685289&0.0000000\\ 
\hline 
00101&0.0157356&1.9685289&0.0000000\\ 
\hline 
00110&0.0157356&1.9685289&0.0000000\\ 
\hline 
00111&0.0157356&1.9685289&0.0000000\\ 
\hline 
01000&0.0157356&1.9554551&0.0000000\\ 
\hline 
01001&0.0157356&1.9554551&0.0000000\\ 
\hline 
01010&0.0157356&1.9554583&0.0000000\\ 
\hline 
01011&0.0157356&1.9554583&0.0000000\\ 
\hline 
01100&0.0157356&1.9554551&0.0000000\\ 
\hline 
01101&0.0157356&1.9554551&0.0000000\\ 
\hline 
01110&0.0157356&1.9554583&0.0000000\\ 
\hline 
01111&0.0157356&1.9554583&0.0000000\\ 
\hline 
10000&0.5898568&0.6338138&0.2320315\\ 
\hline 
10001&0.5898568&0.6336678&0.2320315\\ 
\hline 
10010&0.5895172&0.6343928&0.2281325\\ 
\hline 
10011&0.5895172&0.6342467&0.2298608\\ 
\hline 
10100&0.5898568&0.6338138&0.2320315\\ 
\hline 
10101&0.5898568&0.6336678&0.2320315\\ 
\hline 
10110&0.5895172&0.6343928&0.2281325\\ 
\hline 
10111&0.5895172&0.6342467&0.2298608\\ 
\hline 
11000&0.5969887&0.6236647&0.2343794\\ 
\hline 
11001&0.5969887&0.6235338&0.2350431\\ 
\hline 
11010&0.5967623&0.6237590&0.2262164\\ 
\hline 
11011&0.5967623&0.6236281&0.2262164\\ 
\hline 
11100&0.5969887&0.6236647&0.2343794\\ 
\hline 
11101&0.5969887&0.6235338&0.2350431\\ 
\hline 
11110&0.5967623&0.6237590&0.2262164\\ 
\hline 
11111&0.5967623&0.6236281&0.2262164\\ 
\hline 
\end{tabular}\\
a) & b)
\end{tabular}
\caption{\label{table:metrics_varying_rules_dstc3}Metrics of YARBUS with various rule sets on a) the dstc2\_test and b) the dstc3\_test datasets. The meaning of the rule set number is defined in section~\ref{subsec:varying_rules}. The trackers were run on the SJTU + sys SLU for the DSTC2 dataset and SJTU + err-tied SLU for the DSTC3 dataset.}
\end{table}

\end{document}